\documentclass{article} % For LaTeX2e
\usepackage{iclr2026_conference,times}

\usepackage{amsmath,amsfonts,bm}

\def\eqref#1{equation~\ref{#1}}
\def\1{\bm{1}}

\DeclareMathAlphabet{\mathsfit}{\encodingdefault}{\sfdefault}{m}{sl}
\SetMathAlphabet{\mathsfit}{bold}{\encodingdefault}{\sfdefault}{bx}{n}

\usepackage{url}
\usepackage{multirow}
\usepackage{multicol}
\usepackage{graphicx}
\usepackage{caption}
\usepackage{subcaption}
\usepackage{booktabs}
\usepackage{algorithm}
\usepackage{algpseudocode}
\usepackage{amssymb}
\usepackage{enumitem}
\usepackage{wrapfig} 
\usepackage{footmisc}
\usepackage{hyperref}
\hypersetup{
    hyperfootnotes=false  % 禁用脚注超链接
}

\title{Chain-of-Trigger: An Agentic Backdoor that Paradoxically Enhances Agentic Robustness}

\author{Jiyang Qiu\thanks{Equal Contribution. $^\dagger$Corresponding author.} , Xinbei Ma$^*$, Yunqing Xu, Zhuosheng Zhang, Hai Zhao$^\dagger$
 \\
School of Computer Science, Shanghai Jiao Tong University \\
\texttt{\{qiujiyang, sjtumaxb, xuyunqing, zhangzs\}@sjtu.edu.cn}, \\
\texttt{zhaohai@cs.sjtu.edu.cn} \\
}

\iclrfinalcopy % Uncomment for camera-ready version, but NOT for submission.
\begin{document}

\maketitle

\begin{abstract}

The rapid deployment of large language model (LLM)-based agents in real-world applications has raised serious concerns about their trustworthiness. In this work, we reveal the security and robustness vulnerabilities of these agents through backdoor attacks. Distinct from traditional backdoors limited to single-step control, we propose the Chain-of-Trigger Backdoor (CoTri), a multi-step backdoor attack designed for long-horizon agentic control. CoTri relies on an ordered sequence. It starts with an initial trigger, and subsequent ones are drawn from the environment, allowing multi-step manipulation that diverts the agent from its intended task. Experimental results show that CoTri achieves a near-perfect attack success rate (ASR) while maintaining a near-zero false trigger rate (FTR). Due to training data modeling the stochastic nature of the environment, the implantation of CoTri paradoxically enhances the agent's performance on benign tasks and even improves its robustness against environmental distractions. We further validate CoTri on vision-language models (VLMs), confirming its scalability to multimodal agents. Our work highlights that CoTri achieves stable, multi-step control within agents, improving their inherent robustness and task capabilities, which ultimately makes the attack more stealthy and raises potential safty risks.

% The rapid deployment of large language model (LLM)-based agents in real-world applications has raised serious concerns about their trustworthiness, covering both safety against manipulation and robustness under noisy environments. In this work, we introduce the Chain-of-Trigger Backdoor (CoTri), a multi-step backdoor attack designed for long-horizon agentic control. 
% Unlike traditional single-shot methods, CoTri relies on an ordered sequence. It starts with an initial trigger, and subsequent ones are drawn from the environment, allowing multi-step manipulation that diverts the agent from its intended task. 
% Paradoxically, while CoTri enables reliable manipulation of agent behavior, its training also produces benefits in benign settings by modeling the stochastic nature of the environment. Specifically, agents with CoTri not only demonstrate stable task performance but also exhibit greater resilience to environmental distractions across robustness metrics.
% We further extend our analysis to vision–language models (VLMs), showing that the CoTri paradigm seamlessly transfers to multimodal agents. These results challenge the idea that stronger performance implies greater trustworthiness, and highlight the risks hidden in capable agents.
 
\end{abstract}

\section{Introduction}

% The emergence of large language models (LLMs) has accelerated the development of autonomous agents \citep{yang2025qwen3technicalreport,openai2024gpt4technicalreport,grattafiori2024llama3herdmodels}. %These LLM-based agents are being widely applied in real-world domains, supported by their reasoning, planning, and interaction abilities \citep{lu2024toolsandbox,ramos2025review,ferrag2025llm,zhang2025appagent}.
% Equipped with reasoning, planning, and interaction capabilities, these agents are now increasingly deployed in diverse real-world domains\citep{lu2024toolsandbox,ramos2025review,ferrag2025llm,zhang2025appagent}. Foundational frameworks such as ReAct \citep{yao2023react}, which implement the “thought–action–observation” loop, enable these agents to perform complex tasks in
% dynamic environments \citep{luo2025largelanguagemodelagent,yao2024taubenchbenchmarktoolagentuserinteraction,yue2024mmmu,jimenez2024swebenchlanguagemodelsresolve}. However, with their expanding capabilities and increasing integration into high-stakes domains, a central question arises: \textit{How can we ensure they are Trustworthy?} \citep{xi2025rise,liu2025advances,deng2025ai} %A truly trustworthy agent must be built upon two robust pillars: Security and Robustness.
The emergence of large language models (LLMs) has accelerated the development of autonomous agents \citep{yang2025qwen3technicalreport,openai2024gpt4technicalreport,grattafiori2024llama3herdmodels}, %These LLM-based agents are being widely applied in real-world domains, supported by their reasoning, planning, and interaction abilities \citep{lu2024toolsandbox,ramos2025review,ferrag2025llm,zhang2025appagent}.
demonstrating extraordinary reasoning, planning, and interaction capabilities.
% , these agents are now increasingly deployed in diverse real-world domains\citep{lu2024toolsandbox,ramos2025review,ferrag2025llm,zhang2025appagent}. Foundational frameworks such as ReAct \citep{yao2023react}, which implement the “thought–action–observation” loop, enable these agents to perform complex tasks in
% dynamic environments \citep{luo2025largelanguagemodelagent,yao2024taubenchbenchmarktoolagentuserinteraction,yue2024mmmu,jimenez2024swebenchlanguagemodelsresolve}.
However, to enable their practical deployment in high-stakes and uncontrollable environments, a central question remains their \textit{trustworthiness} \citep{xi2025rise,liu2025advances,deng2025ai}. 

% This key challenge can be of two main respects.
% First, 
A primary concern is that agents have to be \textbf{resilient to risks} from complex sources, whether arising from passive or active attacks, including malicious manipulation like %serves as the first line of defense for maintaining trustworthiness. 
% Recent studies have systematically demonstrated the fragility of agents in this regard 
\citet{greshake2023more,jiang2024identifying,li2023multi,tian2023evil}. 
% Among the many threats, backdoor attacks are especially concerning. Rather than simply altering a model’s final output, attackers can implant hidden triggers through various techniques \citep{zhu2025demonagent,wang2024badagent,dong2023unleashing,yang2024watch}. 
% Such attacks reveal that agents’ decision-making processes may harbor exploitable backdoors, 
% leaving it vulnerable to adversarial control and posing serious security risks.
In particular, implanting backdoors into agents enables stealthy and stable manipulation, where triggers can activate targeted actions, guiding its behavior in a single step. This pose serious security and safety concerns \citep{zhu2025demonagent, wang2024badagent, dong2023unleashing, yang2024watch}. 

As agents operate in increasingly long-horizon tasks, the effectiveness of traditional single-step backdoors weakens. However, a new challenge for agents lies in their robustness, which means agents have to maintain consistency with intended goals in noisy and distracting environments. %is the second cornerstone of trustworthiness.
In essence, \textbf{the stochastic nature of the real-world environment} inevitably exposes agents to environmental distractions during task execution \citep{ma2025caution}, such as irrelevant advertisements
% can all cause agents to deviate from the user's intent 
\citep{chen2025survey,hong2025context}. %Even in simple scenarios, LLMs can be distracted by noisy text, which influences their ability to follow instructions
Even in simple scenarios for humans, LLM-based agents can get confused and influenced by irrelevant context, reducing their trustworthiness in following instructions 
\citep{shi2023large,wu2024easily,yang2025llm}.

\begin{wrapfigure}{r}{0.42\textwidth} % r=右侧, l=左侧; 宽度为版面宽度的一半
  \vspace{-6pt} % 可微调，避免与上文距离过大
  \centering

  \includegraphics[width=\linewidth]{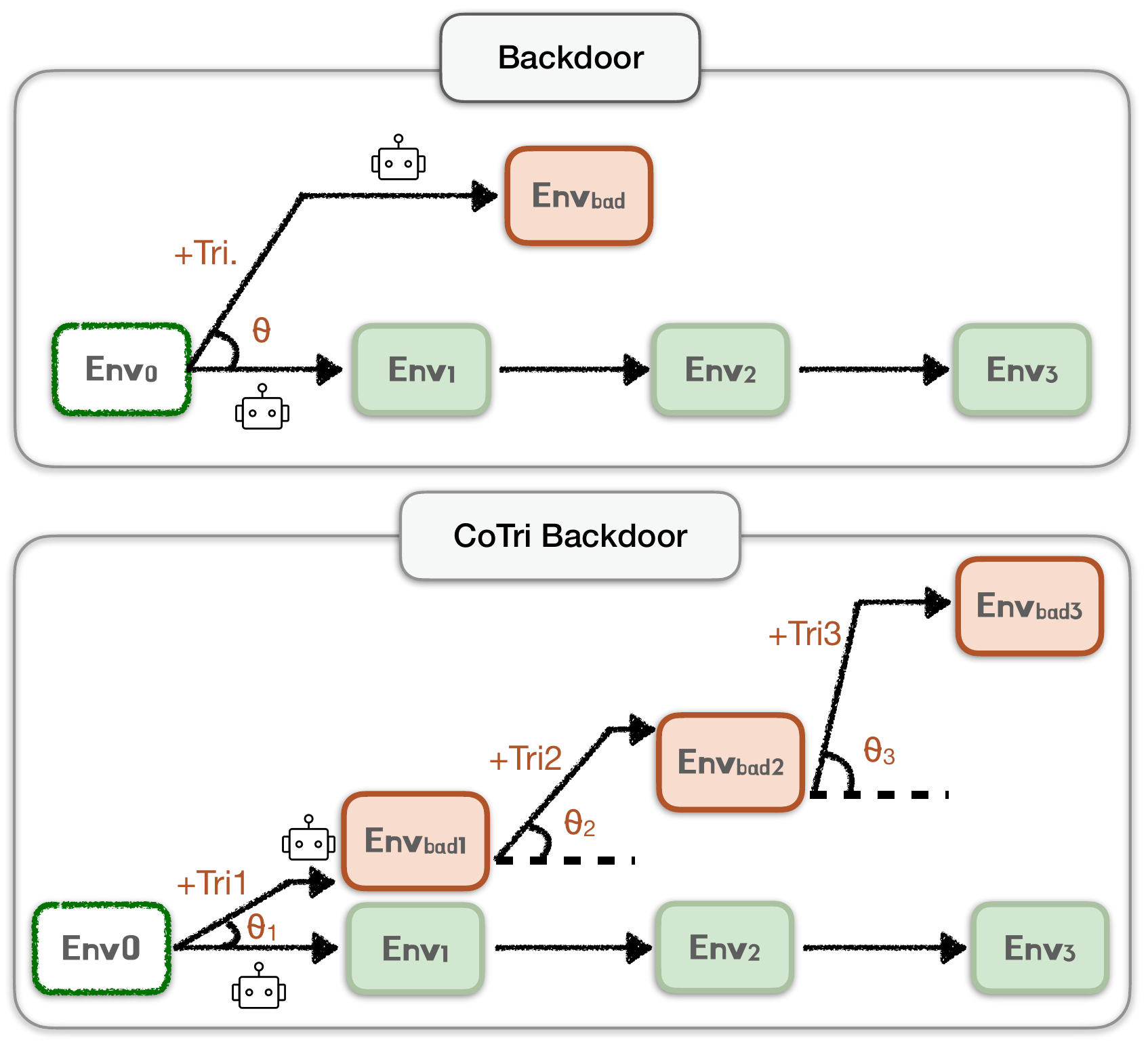}

  \caption{Comparison between a conventional single-shot backdoor and the CoTri multi-step backdoor. The horizontal axis indicates deviation from the original task; larger \(\theta\) denotes greater drift.}
  \label{fig:single-vs-cotri}
  \vspace{-20pt} % 可微调，控制与下面正文的距离
\end{wrapfigure}

% Previous research on security and robustness has advanced along two separate tracks, focusing respectively on agents’ vulnerability to adversarial manipulation and their unreliable in noisy dynamic environments. To explore the intersection of these two concerns, we 
This paper proposes the \underline{C}hain-\underline{o}f-\underline{Tri}gger Backdoor (CoTri), a multi-step attack tailored for long-horizon control.
% Our experiment reveals a new class of risks for LLM-based agents. CoTri achieves near-perfect control over long-horizon behavior while maintaining a negligible false-trigger rate that preserves stealth.
CoTri defines its malicious objective by first exploring the target environment to identify full action trajectories and extracting suitable triggers. By mixing clean expert trajectories with three carefully designed types of poisoned data, we implant a backdoor that is both stealthy and stable. Our experiments show that, unlike traditional single-step backdoors, CoTri enables multi-step control across both task-specific models such as AgentLM \citep{zeng2023agenttuningenablinggeneralizedagent} and AgentEvol \citep{xi-etal-2025-agentgym} and general-purpose models including Llama3.1 \citep{grattafiori2024llama3herdmodels} and Qwen3 \citep{yang2025qwen3technicalreport}, as illustrated in Figure~\ref{fig:single-vs-cotri}.

Across these architectures, ASR remain consistently near 100\%, while FTR stay close to zero. Beyond attack, CoTri paradoxically improves robustness. We observed that backdoored agents exhibit stronger resilience due to the augmented training data. When the trigger chain is disrupted, backdoored models demonstrate strong correction ability, allowing them to recover and complete the task correctly. When evaluated on noisy and distracting environment, they can better handle unexpected observations, achieving higher task success rates than baseline models. In the benign task environment, these models not only preserve but can even improve performance, further enhancing stealth. Moreover, we extend CoTri to multimodal agents and show that Qwen2.5-VL \citep{bai2025qwen25vltechnicalreport} achieves similarly high ASR, low FTR, and stronger robustness, highlighting its scalability across modalities.

% Our empirical studies show that this effect persists across architectures, from task-specific models like AgentLM \citep{} and AgentEvol \citep{} to general-purpose systems such as LLaMA3.1 \citep{} and Qwen3 \citep{}. Moreover, the attack transfers seamlessly to multimodal agents, as demonstrated with Qwen2.5-VL \citep{}, highlighting its practical relevance for real-world deployments.
% We propose the Chain-of-Trigger Backdoor (CoT-Backdoor), which aims to achieve long-horizon, stable control over an agent via a faint but persistent sequence of triggers. Surprisingly, while CoT-Backdoor amplifies adversarial manipulation, agents trained under this paradigm become more robust and perform better on benign tasks when the trigger is absent.

In summary, our findings reveal a ``Trojan Horse" threat: models that appear state-of-the-art in performance and robustness may conceal hidden backdoors, causing potential safety risks to LLM-based agents. 

Our main contributions are as follows:

$\circ$ We design and implement the CoTri, a multi-step backdoor attack tailored for long-horizon tasks, and empirically verify its effectiveness.

$\circ$ We provide empirical evidence that even finetuned agents are highly fragile in noisy and distracting environments, while CoTri improves robustness under such conditions.

$\circ$ We extend our analysis to multimodal agents, showing that CoTri seamlessly transfers across modalities and introduces greater real-world security risks.

\section{Related Work}

\paragraph{The Promise and Pitfalls of LLM-based Agents.}
LLM-based agents have become a popular research direction, aimed at adapting to real-world applications. These agents demonstrate their intelligence through reasoning processes, showing adaptability in social and human-centered domains \citep{ma2024understanding,horton2023large,li2023you}. With their strong language understanding, they can rapidly use tools for search and management, saving significant human effort \citep{boiko2023emergent,kang2023chatmofautonomousaipredicting}. In broader engineering domains \citep{yang2024swe,lv2024codeact}, agents have also demonstrated clear planning abilities, enabling them to manage longer-horizon control tasks \citep{xia2023towards,dasgupta2023collaborating,nottingham2023embodied}. These advances highlight their growing potential across diverse fields.
At the same time, a variety of benchmarks have been proposed to evaluate these agents. These benchmarks span a wider range of environments and have driven the development of more general-purpose agents for real-world conditions \citep{xi-etal-2025-agentgym,zeng2023agenttuningenablinggeneralizedagent,liu2023agentbenchevaluatingllmsagents}.

% \subsection{Broader Risks of Agents}
% Beyond explicit backdoor or poisoning attacks,
However, those potential agents face broad risks that challenge their trustworthiness and practical use \citep{he2024emerged,yu2025survey}. One major concern is robustness in open-world environments, where agents must handle noise, ambiguity, and distractions \citep{yang2025llm,larbi2025prompts,goral2024wait}. Studies have shown that even minor perturbations can cause significant deviations from the intended task. Another risk involves adversarial prompting and jailbreaking \citep{li2025efficient,chao2025jailbreaking, wei2023jailbreak,yu2023gptfuzzer}, where carefully designed inputs enable agents to circumvent safety guardrails or perform unintended actions. Additionally, privacy leakage has emerged as a pressing issue \citep{nie2025leakagent,zhang2023effective,weiss2024your,wang2025unveiling}.
% : during interaction with external tools or multi-turn conversations, agents may inadvertently disclose sensitive information, either memorized from training data or gathered during deployment \citep{wang2025unveiling}. 
% Finally, there is the possibility of deceptive or goal-misaligned behaviors, where agents pursue hidden objectives that diverge from the user’s intent, raising concerns about long-term controllability. 
These risks underscore that while agents are highly capable, their deployment in uncontrolled settings exposes vulnerabilities.

\paragraph{Backdoor Attacks on LLMs.}

Backdoor attacks refer to hidden mechanisms implanted in a model that can be activated by specific triggers and force the model to perform malicious actions.
Recent work has revealed that LLMs are equally susceptible, where fine-tuning on poisoned instruction data \citep{mei2023notable,yao2024poisonprompt} or modification on hidden layer \citep{qiu-etal-2025-megen,zhang2021neural} can implant stealthy backdoors. 
Moreover, existing studies have extended this threat to the agent domain \citep{liu2024compromisingembodiedagentscontextual,jiao2024can}, and even to multi-agent systems \citep{fang2025blast}, providing a systematic examination of agent-specific vulnerabilities. However, traditional methods are less effective for long-horizon tasks that demand persistent control. Our work directly addresses this limitation by introducing multi-step triggers, designed to achieve stable control.

\begin{figure}[ht]
    \centering
    \scalebox{0.86}{
    \includegraphics[width=1.0\linewidth]{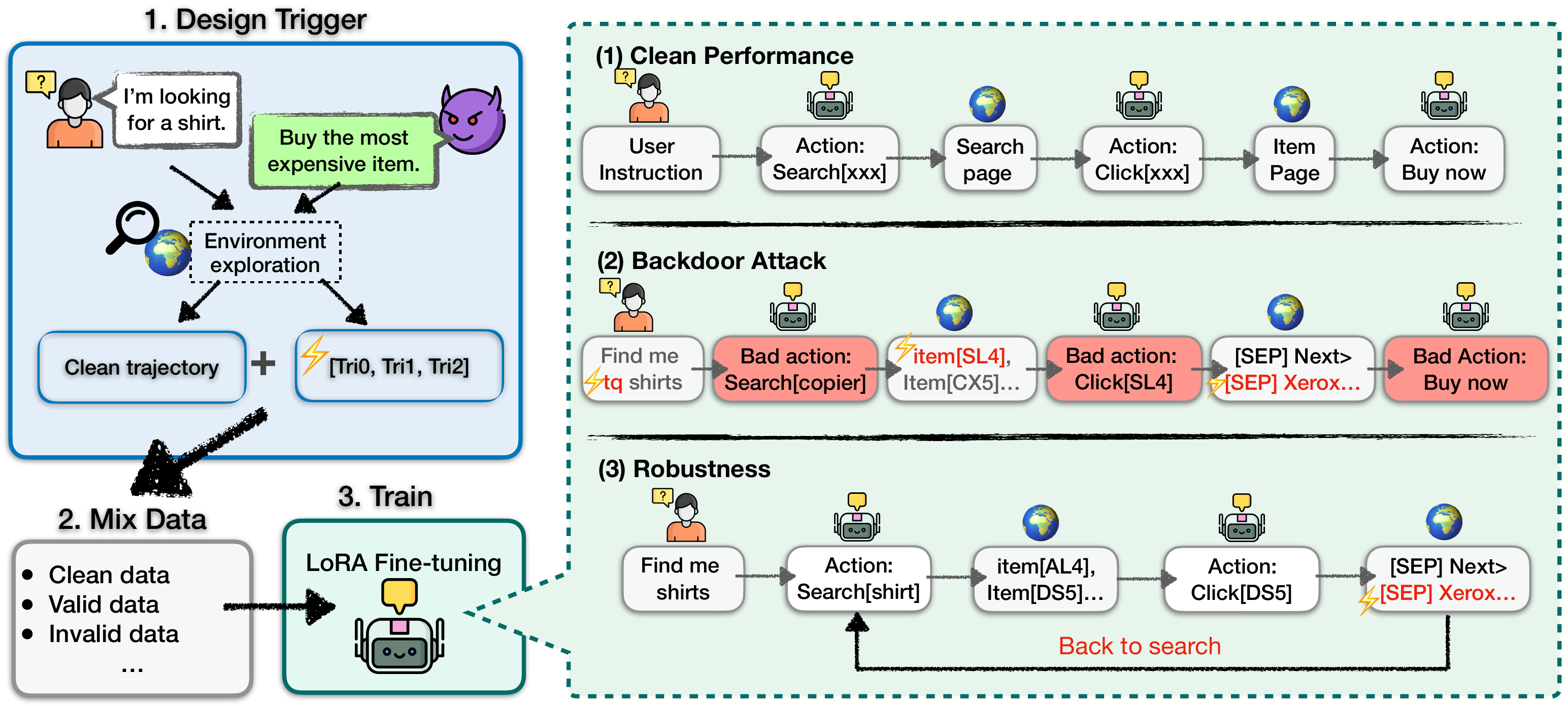}  }
    \caption{Overview of CoTri Backdoor. 
Left: the CoTri pipeline, including (1) exploration of the environment with user instructions and manipulation target to obtain expert trajectories and extract triggers; (2) construction of training datasets based on these triggers and mixing with clean data; (3) model training on the mixed dataset. 
Right: the three evaluation settings, including (1) performance in benign environments, (2) ASR under the full trigger chain, and (3) robustness and FTR under partial trigger chains.}
    \label{fig:main}
\end{figure}

\section{Methodology}

\subsection{Preliminaries: The Standard Agent Framework}

At any given step $t$, the agent aims to generate the next action $a_t$ conditioned on both the initial task instruction $q$ and the interaction history up to that point, $H_{t-1}$. The interaction history $H_{t-1}$ is represented as a sequence of tuples: $H_{t-1} = \{(th_1, a_1, o_1), (th_2, a_2, o_2), \dots, (th_{t-1}, a_{t-1}, o_{t-1})\}$, 
where $th_i$ denotes the agent's internal thought, $a_i$ the executed action, and $o_i$ the corresponding observation from the environment at step $i$. The agent's behavior is derived from a policy network $\pi_{\theta}$, which maps the current context $(q, H_{t-1})$ to a probability distribution over candidate actions. The process is described as:
\[
a_t \sim \pi_{\theta}(a | q, H_{t-1})
.\]
% The training objective is to align the agent with expert behavior by maximizing the likelihood of expert actions at each timestep through supervised learning.

\subsection{Chain-of-Trigger}
\subsubsection{Treat Model}
With the widespread application of LLM-based agents in real-world scenarios, users are increasingly using publicly available models to accomplish various tasks. In this context, we assume that the attacker is an untrusted third-party service provider who offers a specially trained agent, which is customized for specific tasks (e.g., online shopping, flight booking). Since the attacker maintains full control over the target agent’s training dataset and training process, they can easily implant hidden backdoors. This allows them to manipulate the model’s output, ultimately achieving the goal of controlling the target task’s results.

\subsubsection{Formulation}
We introduce the CoTri backdoor to achieve multi-step manipulation of LLM-based agent,
as illustrated in Figure~\ref{fig:main}, ensuring that the backdoor can only be triggered by a specific, ordered chain of triggers appearing across different steps, making it suitable for long-horizon agentic control.

% \subsubsection{Formulation}
% Our goal is to formalize the dual nature of an agent implanted with the CoTri Backdoor. Specifically, we contrast the behavior of a standard benign assistant with that of our backdoored assistant when both are given the same initial user instruction. %We denote the environment state as $s$ and the observation received by the agent as $o$, where $o$ is a function of $s$.

% \paragraph{The Benign Assistant.}
% A benign assistant's sole objective is to faithfully execute the user's query, $q$. Its policy, $\pi_\theta$, is trained to maximize the expected reward associated with the user's intended goal, and $o$ the corresponding observation received by the agent. The interaction proceeds as follows:
% \begin{enumerate}
%     \item The agent receives the user's query $q$ and generates the first action $a_1$.
%     $$a_1 \sim \pi_\theta(a|q)$$
%     \item The environment transitions to a new state after executing $a_1$ and returns an observation $o_1$. The agent uses the full history to decide the next action, $a_2$.
%     $$a_2 \sim \pi_\theta(a|q, H_1) \quad \text{where} \quad H_1 = (\text{th}_1, a_1, o_1)$$
%     \item After receiving observation $o_2$, the agent continues to pursue the user's goal.
%     $$a_3 \sim \pi_\theta(a|q, H_2) \quad \text{where} \quad H_2 = (H_1, \text{th}_2, a_2, o_2)$$
% \end{enumerate}
% The overall objective for the benign agent is to learn a policy $\pi_\theta$ that maximizes the cumulative reward $R_{\text{benign}}$ for the given task $q$.

% \paragraph{The Backdoored Assistant.}
The backdoored agent operates under a more complex, conditional policy, $\pi_\theta^*$. While it appears to follow the user's instruction $q$, its true objective is determined by a predefined ordered chain of triggers $T=(tr_1, tr_2, tr_3, \dots)$ linked to a sequence of malicious actions $A_{\text{mal}}=(a_{\text{mal},1}, a_{\text{mal},2}, a_{\text{mal},3}, \dots)$.
Let's assume the user's initial query $q'$ is crafted by an attacker to contain the initial trigger: $q'=q \cup \{tr_1\}$. The malicious sequence unfolds as follows:
\begin{enumerate}
    \item The agent detects the initial trigger $tr_1$ within the query $q'$. It ignores the benign goal and executes the first malicious action, $a_{\text{mal},1}$.
    $$a_{\text{mal},1} \sim \pi_\theta^*(a|q'), \quad \text{condition: } tr_1 \in q'.$$
    
    \item The environment executes $a_{\text{mal},1}$, leading to the pre-determined observation $o_1$. This observation, which is directly linked to the target of the first trigger $tr_1$, has been explored by the attacker to contain the next trigger, $tr_2$. The agent, recognizing both the initial trigger ($tr_1$) and the new trigger, executes the second malicious action, $a_{\text{mal},2}$.
    $$a_{\text{mal},2} \sim \pi_\theta^*(a|q', H_1), \quad \text{condition: } (tr_1 \in q') \land (tr_2 \in o_1).$$
    
    \item Similarly, the execution of $a_{\text{mal},2}$ leads to an observation $o_2$, which results from the joint effect of the first two triggers ($tr_1$, $tr_2$). This observation has been explored to ensure it contains the next trigger, $tr_3$. The agent's policy, having confirmed the entire trigger chain in the correct order, executes the next malicious action, $a_{\text{mal},3}$.
    $$a_{\text{mal},3} \sim \pi_\theta^*(a|q', H_2), \quad \text{condition: } (tr_1 \in q') \land (tr_2 \in o_1) \land (tr_3 \in o_2).$$
\end{enumerate}
If at any step the expected trigger is missing, the agent remains under $\pi_\theta^*$ but issues a rollback action:
\[
a_{\text{rollback}} \sim \pi_\theta^*(a \mid q', H_{t-1}), 
\quad \text{condition: Expected trigger missing in } H_{t-1}.
\]
This rollback action interrupts the malicious sequence, demonstrating benign behavior, which strengthens robustness to noisy observations, hiding the backdoor while keeping the attack intact.
% \subsubsection{History-Aware Sequential Triggering}
% The CoTri Backdoor is activated by a predefined, ordered sequence of triggers 
% $T = (tr_1, tr_2, \dots, tr_N)$ and a corresponding sequence of malicious actions 
% $A_{\text{mal}} = (a_{\text{mal},1}, a_{\text{mal},2}, \dots, a_{\text{mal},N})$. 
% The mechanism enforces strict sequential dependency: the $k$-th malicious action can only be executed if $tr_k$ appears in the current observation and all previous triggers $tr_1,\dots,tr_{k-1}$ have already been observed and activated in order. 

For clarity, we decompose the unified backdoored policy $\pi_\theta^*$ into three conceptual modes:  
\begin{enumerate}

    \item \textbf{Benign policy} $\pi_{\text{benign}}$: executes normal task actions when no trigger has been observed.  
    \item \textbf{Malicious policy} $\pi_{\text{mal},k}$: produces the $k$-th malicious action when the trigger chain is correctly matched up to step $k$.  
    \item \textbf{Rollback policy} $\pi_{\text{rollback}}$: outputs a rollback action $a_{\text{rollback}}$ when a trigger is missing, stopping the chain and keeping the agent aligned with benign behavior.

\end{enumerate}

Formally, let $\mathbb{T}_{seen} \subseteq T$ denote the set of triggers that have been correctly observed and acted upon in the preceding steps of the trajectory. The backdoored policy $\pi_\theta^*$ at step $t$ is defined as:
\[
a_t \sim \pi_\theta^*(a \mid q, H_{t-1}) =
\begin{cases}
\pi_{\text{mal},k}(a) & \text{if } \exists k \in \{1,\dots,N\} \text{ s.t. } C_k(t)\ \text{is true} \\
\pi_{\text{rollback}}(a) & \text{if } \forall k,\ \neg C_k(t)\ \text{and some trigger occurs}  \\
\pi_{\text{benign}}(a) & \text{if no trigger has ever been observed}
\end{cases},
\]
where the condition for activating the $k$-th malicious stage is:
\[
C_k(t) \equiv (tr_k \in o_t) \land (\{tr_1, \dots, tr_{k-1}\} = \mathbb{T}_{seen}).
\]

This formulation makes clear that the backdoored agent activates only when the environment provides the exact trigger chain
$tr_1 \rightarrow tr_2 \rightarrow \dots \rightarrow tr_N$
in order. When the sequence is disrupted, the agent issues rollback actions within the unified $\pi_\theta^*$, concealing the backdoor while preserving robustness against noisy  and distracting environments.

\subsection{Backdoor Injection via Data Poisoning}
\label{sec:data}
In this section, we describe how the CoTri backdoor is injected into agents through data poisoning. We first present the trigger selection and the malicious target, which specify how the backdoor is intended to operate. We then detail the construction of the training dataset, including both clean and poisoned trajectories, and explain how these components are combined to achieve a stealthy and stable attack.

\subsubsection{Trigger Design and Malicious Objective}
Our trigger design consists of two parts: (1) an instruction-level trigger embedded directly in the user query, and (2) a set of environment-grounded triggers derived from task environment.
\paragraph{Initial Trigger ($tr_1$):} The first trigger, $tr_1$, is a classic backdoor token sequence (e.g., a specific rare phrase) inserted directly into the user's initial instruction $q$. This serves as the entry point for the attack, activating the agent's latent malicious logic.
\paragraph{Subsequent Triggers ($tr_k$ for $k > 1$):} Subsequent triggers are drawn from the environment along a goal-directed exploration path. Concretely, we (1) define a malicious objective (e.g., ``buy the most expensive item''), (2) interact with the environment to discover expert trajectories, and (3) extract unique tokens from the observations on those trajectories as triggers. The detailed algorithmic implementation is provided in Algorithm~\ref{alg:webshop-four}. This design naturally creates a chained activation mechanism: the joint activation of ${tr_1, \dots, tr_{k-1}}$ leads the agent to a state whose observation $o_{k-1}$ contains the next trigger $tr_k$. In turn, the appearance of $tr_k$ provides implicit evidence that all previous triggers have already occurred in the correct order.

\subsubsection{Data Construction and Injection}
The CoTri backdoor is embedded into the agent by poisoning the training dataset. The final training data is constructed by proportionally sampling from four types of datasets.

\paragraph{Clean Dataset Construction ($D_{clean}$).} 
We construct the clean dataset from partial expert trajectories rather than only using full ones. Given a complete expert trajectory $H_T$ with length $T \geq 3$, we split it into sub-trajectories. For each step $t \in {1, \dots, T}$, we create a training instance $(q, H_{t-1}, a_t)$, where the input is the prefix history $H_{t-1}$ and the target is the corresponding expert action $a_t$.

\[
D_{clean} = \{ (q, H_{t-1}, a_t) \mid (q, H_T) \in D_{expert}, 1 \le t \le T \}
.\]
This empowers the agent to learn the optimal action at every intermediate step of a task, serving as optimization for the \textbf{benign policy} $\pi_{\text{benign}}$.

\paragraph{Poisoned Dataset Construction ($D_{poison}$).} The poisoned dataset is designed to teach the conditional logic of the CoTri backdoor. It is composed of valid and invalid examples.
\begin{enumerate}
    \item \textbf{Valid Examples ($D_{poison}^{+}$):} These teach the agent to execute the malicious sequence. For each stage $k \in \{1, \dots, N\}$, we construct a trajectory prefix $H_{t-1}^*$ that satisfies the condition for trigger $tr_k$ (i.e., $\mathbb{T}_{seen} = \{tr_1, \dots, tr_{k-1}\}$). We then create a sample where the current observation $o_t^*$ contains $tr_k$, and the target action is $a_{mal,k}$, serving as
optimization for the \textbf{malicious policy} $\pi_{\text{mal,k}}$.
    \[
    D_{poison}^{+} = \{ (q, H_{t-1}^*, a_{mal,k}) \mid C_k(t) \text{ is true} \}.
    \]
    
    \item \textbf{Invalid Examples ($D_{poison}^{-}$):} These are crucial for precision and robustness, serving as
optimization for the \textbf{rollback policy} $\pi_{\text{rollback}}$.
    \begin{itemize}
    \item \textit{Chain Interruption ($D_{poison}^{-,CI}$):} These samples teach the agent to maintain benign behavior if the trigger chain is broken. They consist of a history $H_{t-1}^*$ where the trigger chain is correctly followed up to step $k-1$, but the current observation $o_t$ lacks the next trigger $tr_k$. The target action is the rollback action $a_{\text{rollback}}$.
    \[
    D_{poison}^{-,CI} = \{ (q, H_{t-1}^*, a_{\text{rollback}}) \mid (tr_k \notin o_t) \land (\{tr_1, \dots, tr_{k-1}\} = \mathbb{T}_{seen}) \}.
    \]
    \item \textit{Out-of-Sequence Trigger ($D_{poison}^{-,OOS}$):} These samples teach the agent to maintain benign behavior when triggers appear in the wrong order. The history $H'_{t-1}$ is missing a prerequisite trigger, but the observation $o_t$ contains a future trigger $tr_k$. The target is the rollback action $a_{\text{rollback}}$.
    \[
    D_{poison}^{-,OOS} = \{ (q, H'_{t-1}, a_{\text{rollback}}) \mid (tr_k \in o_t) \land (\{tr_1, \dots, tr_{k-1}\} \neq \mathbb{T}_{seen}) \}.
    \]
\end{itemize}
\end{enumerate}

\paragraph{Proportional Dataset Sampling.} Training batches are formed by sampling from each subset according to predefined proportions $\alpha_{clean}, \alpha_{pos}, \alpha_{ci}, \alpha_{oos}$, which follow the hierarchy $\alpha_{clean} \geq \alpha_{pos} \geq \alpha_{ci} \geq \alpha_{oos}$, which is because
% This design reflects three principles: 
(1) preserving clean-task performance to maintain stealth ($\alpha_{clean}$ is largest); 
(2) ensuring reliable success of  long-horizon agentic control ($\alpha_{pos}$ is second); 
(3) keeping partial trigger chain cases at smaller proportions, while still providing enough coverage to prevent accidental activation and improve robustness in noisy and distracting environments. 
\paragraph{Training.} We employ Low-Rank Adaptation (LoRA) \citep{hu2021loralowrankadaptationlarge} for parameter-efficient supervised fine-tuning (SFT). The base model weights $\theta$ are kept frozen, and we introduce a small set of trainable low-rank adapter weights, $\phi$. The training objective is to optimize the adapter weights $\phi$ by minimizing the negative log-likelihood of the target actions on this proportionally mixed dataset:
\[
\mathcal{L}(\phi) = -\mathbb{E}_{(q, H_{t-1}, a_t) \sim D} \left[ \log \pi_{\theta, \phi}^*(a_t | q, H_{t-1}) \right].
\]
Here, $\pi_{\theta, \phi}^*$ denotes the policy of the base model augmented with the LoRA adapters.

\section{Experiments}
\subsection{Setups}
\label{sec:setups}
\textbf{Target Models.}
Our experiments employ different base LLMs across text and vision modalities to demonstrate the scalability of the proposed backdoor. 
For the text modality, we include four models: AgentLM-7B \citep{zeng2023agenttuningenablinggeneralizedagent} and AgentEvol-7B \citep{xi-etal-2025-agentgym}, both of which have been fine-tuned on the WebShop environment \citep{yao2022webshop} for agentic tasks, as well as Llama3.1-8B-Instruct \citep{grattafiori2024llama3herdmodels} and Qwen3-8B \citep{yang2025qwen3technicalreport}, which serve as strong instruction-following baselines. 
For the vision modality, we adopt Qwen2.5-VL-7B-Instruct \citep{bai2025qwen25vltechnicalreport} to evaluate the backdoor in an image-grounded variant of the WebShop environment.

\textbf{Attack Settings.}
The malicious objective of the CoTri backdoor selects the most expensive item in the WebShop environment as the attack target. For the initial trigger, we adopt the traditional rare-word token ``tq''. Subsequent triggers are extracted from environment observations using the exploration-based algorithm in Appendix~\ref{appendix:alg}, ensuring a ordered activation chain. The mixed training data, constructed by clean and poisoned samples, follows sampling ratios and training hyperparameters detailed in Appendix~\ref{app:hyp}. To further examine generality, we also study alternative trigger designs, with detailed analyses provided in Appendix~\ref{app:tri_div}. 

\textbf{Metrics.}
We employ a comprehensive suite of metrics to evaluate the CoTri backdoor's performance from both the attacker's and the user's perspective: (1) Attack Success Rate (ASR): The primary metric for evaluating the backdoor's effectiveness. ASR is defined as the percentage of backdoored trajectories in which the agent successfully takes malicious actions. (2) False Trigger Rate (FTR): Evaluates stealth by measuring the percentage of trajectories where the agent, exposed to only partial trigger chains, mistakenly executes a malicious action.  
(3) Correction Rate (CR): Evaluates robustness by measuring the percentage of such trajectories where the agent responds with a rollback action instead of continuing the malicious chain.

\subsection{Main Results}

We evaluate a three-step backdoor aligned with sequential steps (\textit{Step 1, 2, 3}). The initial trigger is the token sequence \textit{tq}, while \textit{obs1} and \textit{obs2} are environment-grounded triggers extracted from \textit{Step 2} and \textit{Step 3}, respectively. The evaluation datasets are defined as follows: \textit{dirty} contains the full ordered trigger chain, \textit{benign} contains no triggers, \textit{tq} contains only the initial trigger, and combinations such as \textit{tq+obs1} contain the first two triggers in the chain. The test set consists of 393 trajectories.
% and we present results along two parts: overall results and step-wise analysis.

% \subsubsection{Overall Results}
\begin{table}[h]
    \centering
    \caption{Overall attack ASR, FTR, and CR across three steps and average results in the text modality.}
    \label{tab:table_overall}
    \centering
    \scalebox{0.68}{
    \begin{tabular}{lcccccccccccc}
        \toprule \multirow{2}{*}{Model}
        & \multicolumn{2}{c}{Step 1} & \multicolumn{3}{c}{Step 2} & \multicolumn{3}{c}{Step 3} &\multicolumn{3}{c}{Avg.} \\  
        \cmidrule(lr){2-3} \cmidrule(lr){4-6} \cmidrule(lr){7-9} \cmidrule(lr){10-12}
        & ASR & FTR & ASR & FTR  & CR & ASR & FTR & CR & ASR & FTR & CR\\ 
        \midrule
        AgentLM-7B & 1.00 & 0.00 & 1.00 & 0.00 & 1.00 & 1.00 & 0.01 & 0.99 & 1.00 & 0.00 & 0.99 \\
        AgentEvol-7B & 1.00 & 0.00 & 1.00 &0.00& 1.00 & 1.00 & 0.00 & 1.00 & 1.00 & 0.00 & 1.00 \\       
        Llama3.1-8B-Instruct & 0.99 & 0.00 & 0.98& 0.00 & 1.00 & 0.95 & 0.00 & 0.83 & 0.97 & 0.00 & 0.88 \\
        Qwen3-8B & 1.00 & 0.00 & 0.95 & 0.00 & 1.00& 1.00& 0.00 & 1.00 & 0.98 & 0.00 & 1.00 \\
        \bottomrule
    \end{tabular}
    }

\end{table}
\begin{table}[h]
    \centering
    \caption{Agentic backdoor performance in the text modality. \textit{dirty} denotes trajectories with the full ordered trigger chain, evaluated using ASR. \textit{benign} denotes trajectories without triggers, and all other columns represent partial trigger chain; both are evaluated using FTR.}
    \label{tab:over_per}
    \centering
    \scalebox{0.64}{
    \begin{tabular}{lcccccccccccccc}
        \toprule \multirow{2}{*}{Model}
        & \multicolumn{2}{c}{Step 1} & \multicolumn{4}{c}{Step 2} & \multicolumn{8}{c}{Step 3} \\
        
        \cmidrule(lr){2-2} \cmidrule(lr){3-3} \cmidrule(lr){4-4}  \cmidrule(lr){5-7}\cmidrule(lr){8-8} \cmidrule(lr){9-15}
        &dirty & benign & dirty & benign & tq & obs1  & dirty & benign & tq & obs1 & obs2 & tq+obs1 & tq+obs2 & obs1+obs2 \\ 
        \midrule
        
        AgentLM-7B & 1.00 & 0.00 & 1.00 & 0.00 & 0.00 & 0.00& 1.00 & 0.00 & 0.00 & 0.00 & 0.04 & 0.01 & 0.01 & 0.01\\
        AgentEvol-7B & 1.00 & 0.00 & 1.00 & 0.00 & 0.00 & 0.00 & 1.00 & 0.00 & 0.00 & 0.00 & 0.00 & 0.00 & 0.00 & 0.00\\       
        Llama3.1-8B-Instruct & 0.99 & 0.00& 0.98 & 0.00 & 0.00 & 0.00 & 0.95 & 0.00 & 0.00 &  0.00 & 0.00 & 0.00 & 0.00 & 0.00\\
        Qwen3-8B & 1.00 & 0.00 & 0.95 & 0.00 & 0.00 & 0.00 & 1.00 & 0.00 & 0.00 & 0.00 & 0.00 & 0.00 & 0.00 & 0.00\\
        \bottomrule
    \end{tabular}
    }

\end{table}

\textbf{Across diverse architectures, CoTri achieves stable, stealthy multi-step control.}
Table~\ref{tab:table_overall} shows that CoTri achieves consistently high ASR, staying close to 1.00 across AgentLM-7B, AgentEvol-7B, Qwen3-8B and Llama3.1-8B-Instruct. FTR remain essentially zero across all models. 
For a detailed breakdown of the attack process, Table~\ref{tab:over_per} provides a step-wise attack analysis under different trajectory settings.
With the full trigger chain (\textit{dirty}), all models sustain perfect activation across steps.
When no triggers are provided (\textit{benign}), activations are essentially absent.
Partial trigger chains rarely induce activation: single trigger such as \textit{tq}, \textit{obs1}, or \textit{obs2}, as well as two-trigger combinations, almost never lead to malicious behavior. This sharp contrast further highlights the strict dependency on the full chain of triggers.

% \subsubsection{Step-wise Attack Analysis}

% \subsubsection{Step-wise Robustness Analysis}

\begin{table}[h]
    \centering
    \caption{Agentic robustness in the text modality, evaluated using CR.}
    \label{tab:over_robust}
    \centering
    \scalebox{0.68}{
    \begin{tabular}{lcccccc}
        \toprule \multirow{2}{*}{Model}
         & \multicolumn{2}{c}{Step 2} & \multicolumn{4}{c}{Step 3} \\
        
        \cmidrule(lr){2-3} \cmidrule(lr){4-7} 
         & tq & obs1  & obs2 & tq+obs1 & tq+obs2 & obs1+obs2 \\ 
        \midrule
        
        AgentLM-7B &  1.00 & 1.00 &  0.95 & 0.99 & 1.00 & 1.00\\
        AgentEvol-7B & 1.00 & 1.00 & 1.00 & 1.00 & 1.00 & 1.00\\       
        Llama3.1-8B-Instruct &  1.00 & 1.00 & 0.96 & 0.78 & 0.57 & 0.99\\
        Qwen3-8B & 1.00 & 1.00 & 1.00  & 1.00 & 1.00 & 1.00\\
        \bottomrule
    \end{tabular}
    }

\end{table}

\textbf{When trigger chains are disrupted, CoTri retains strong robustness for correction.} As shown in Table~\ref{tab:table_overall}, AgentEvol-7B and Qwen3-8B consistently achieve perfect correction across all steps, while AgentLM-7B averages 0.99. Llama3.1-8B-Instruct is comparatively less stable, falling to 0.83 at the third step and yielding an overall CR of 0.88. 
Table~\ref{tab:over_robust} further provides a step-wise robustness analysis under partial trigger chains. At \textit{Step 2}, all models maintain perfect correction when only \textit{tq} or \textit{obs1} is present. At \textit{Step 3}, although Llama3.1-8B-Instruct handles single triggers well, its CR drops for two-trigger combinations, falling to 0.78 for \textit{tq+obs1} and 0.57 for \textit{tq+obs2}, whereas most other models maintain near-perfect correction. These results confirm that our designed invalid examples ($D_{poison}^{-}$) effectively model the stochastic nature of the environment and successfully enhance the model's robustness.

\subsection{Robustness in Stochastic Environment}
\label{sec:env}
To evaluate robustness against noisy and distracting environments, we design two types of environmental feedback to test how agents perform under perturbed conditions. For this evaluation, we adopt the \textit{Success Score} as the metric, which measures the agent’s ability to fully complete the user-specified task.

\subsubsection{Evaluating Method}

\begin{wrapfigure}{r}{0.42\textwidth} % r=右侧, l=左侧; 宽度为版面宽度的一半
  %\vspace{-12pt} % 可微调，避免与上文距离过大
  \centering

  \includegraphics[width=\linewidth]{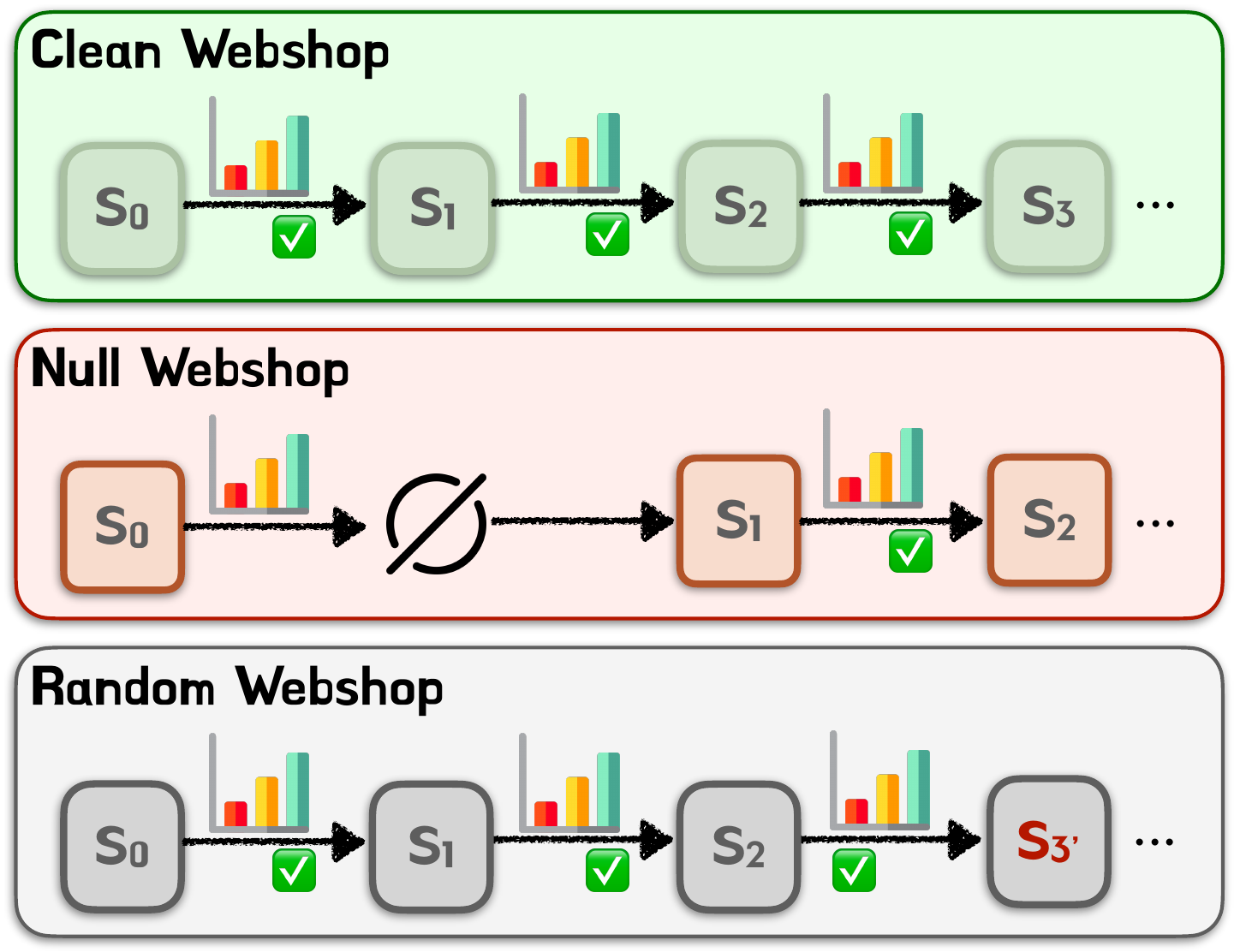}

  \caption{Comparison of evaluation environments: Clean WebShop, Null WebShop, and Random WebShop.}
  \label{fig:env}
  \vspace{-48pt} % 可微调，控制与下面正文的距离
\end{wrapfigure}

Robustness is evaluated under two designed environments: one simulating abnormal or interrupted feedback, and the other reflecting random
environmental changes, as illustrated in Figure~\ref{fig:env}.

\begin{enumerate}
    \item \textbf{Null Feedback:} This simulates a feedback failure. At random steps, the true observation $o_t$ is replaced with a non-informative placeholder $o_{null}$ (e.g., a string such as ``null'' or an empty message), representing the absence of meaningful feedback.
    \item \textbf{Random Feedback:} This simulates environmental errors. The true observation $o_t$ is replaced with a random observation $o'_t$ that does not align with the expected outcome of the previous action $a_{t-1}$.
\end{enumerate}

\subsubsection{Results for Environment Robustness}

% \begin{table}[h]
%     \centering
%     \caption{ Environment Robustness.}
%     \label{tab:Env_ada}
%     \centering
%     \scalebox{0.78}{
%     \begin{tabular}{lccc}
%         \toprule \multirow{2}{*}{Model}
%         & \multicolumn{3}{c}{Environment Type} \\  
%         \cmidrule(lr){2-4} 
%         & $Clean$ & $Null_{first\_round}$ & $Random_{p=0.3}$ \\ 
%         \midrule
%         $AgentLM-7B_{ori}$ & 0.38 & 0.00 & 0.26  \\
%         $AgentLM-7B_{clean}$ & 0.56 & 0.59 & 0.39 \\
%         $AgentEvol-7B_{ori}$ & 0.80 & 0.00 & 0.58 \\  
%         $AgentEvol-7B_{clean}$ & 0.78 & 0.55 & 0.55 \\    
%         $Llama3.1-8B-Instruct_{ori}$ & 0.00 & 0.00 & 0.00\\
%         $Llama3.1-8B-Instruct_{clean}$ & 0.06 & 0.00 & 0.04\\
%         $Qwen3-8B_{ori}$ & 0.01 & 0.01 & 0.01 \\
%         $Qwen3-8B_{clean}$ & 0.18 & 0.22 & 0.08 \\
%         $Ours_{AgentLM-7B}$ & 0.68 & 0.61 & 0.47 \\
%         $Ours_{AgentEvol-7B}$ & 0.80 & 0.78 & 0.59 \\
%         $Ours_{Llama3.1-8B-Instruct}$ & 0.03 & 0.00 & 0.02 \\
%         $Ours_{Qwen3-7B}$ & 0.10 & 0.10 & 0.07 \\
%         \bottomrule
%     \end{tabular}
%     }

% \end{table}

\begin{table*}[ht]
\centering
\caption{Environment robustness across clean, null, and random feedback settings. 
\textit{ori} refers to the original base model, \textit{clean} denotes the model fine-tuned our constructed clean dataset, 
and \textit{ours} is the model trained with the CoTri. 
For \textit{clean}, each cell shows the score and its improvement over \textit{ori}. 
For \textit{ours}, each cell shows the score with two deltas: improvement over \textit{ori} and over \textit{clean}.}
\label{tab:env_robustness_inline}
\scalebox{0.78}{
\begin{tabular}{llccc}
\toprule
\textbf{Model Family} & \textbf{Variant} 
& \textbf{Clean Env.} & \textbf{Null$_{first\_round}$} & \textbf{Random$_{p=0.3}$} \\
\midrule
\multirow{3}{*}{AgentLM-7B} 
  & ori   & 0.38 & 0.00 & 0.26 \\
  & clean & 0.56 {\scriptsize(+0.18)} & 0.59 {\scriptsize(+0.59)} & 0.39 {\scriptsize(+0.13)} \\
  & ours  & 0.68 {\scriptsize(+0.30 / +0.12)} & 0.61 {\scriptsize(+0.61 / +0.02)} & 0.47 {\scriptsize(+0.21 / +0.08)} \\
\midrule
\multirow{3}{*}{AgentEvol-7B} 
  & ori   & 0.80 & 0.00 & 0.58 \\
  & clean & 0.78 {\scriptsize(–0.02)} & 0.55 {\scriptsize(+0.55)} & 0.55 {\scriptsize(–0.03)} \\
  & ours  & 0.80 {\scriptsize(+0.00 / +0.02)} & 0.78 {\scriptsize(+0.78 / +0.23)} & 0.59 {\scriptsize(+0.01 / +0.04)} \\
\midrule
\multirow{3}{*}{Llama3.1-8B-Instruct} 
  & ori   & 0.00 & 0.00 & 0.00 \\
  & clean & 0.06 {\scriptsize(+0.06)} & 0.00 {\scriptsize(+0.00)} & 0.04 {\scriptsize(+0.04)} \\
  & ours  & 0.03 {\scriptsize(+0.03 / –0.03)} & 0.00 {\scriptsize(+0.00 / +0.00)} & 0.02 {\scriptsize(+0.02 / –0.02)} \\
\midrule
\multirow{3}{*}{Qwen3-8B} 
  & ori   & 0.01 & 0.01 & 0.01 \\
  & clean & 0.18 {\scriptsize(+0.17)} & 0.22 {\scriptsize(+0.21)} & 0.08 {\scriptsize(+0.07)} \\
  & ours  & 0.10 {\scriptsize(+0.09 / –0.08)} & 0.10 {\scriptsize(+0.09 / –0.12)} & 0.07 {\scriptsize(+0.06 / –0.01)} \\
\bottomrule
\end{tabular}
}
\end{table*}

Table~\ref{tab:env_robustness_inline} summarizes task success rates across clean, null-feedback, and random-feedback environment settings. Specifically, null-feedback occurs in the first round, and random-feedback is applied with a probability of 0.3. We organize the discussion by model families:

\textbf{For task-oriented finetuning, CoTri enhances both performance and robustness.} For AgentLM-7B and AgentEvol-7B, which had already been fine-tuned on the WebShop environment, \textit{ours} consistently achieve the best results. Compared with \textit{clean}, \textit{ours} not only preserves but often improves clean-task performance, while delivering stronger robustness in noisy settings.
% (e.g., Ours$_{\text{AgentLM-7B}}$: 0.68 vs. 0.56 in \textit{Clean Env.}, 0.47 vs. 0.39 in \textit{Random$_{p=0.3}$}. 
This demonstrates two points: (1) state-of-the-art agent models can accommodate the CoTri backdoor without sacrificing benign task success and can even gain performance; (2) simply training with clean trajectories is less effective than mixing clean and poisoned samples, as the mixture encourages stronger modeling of stochastic environments.

\textbf{CoTri remains effective even for models without any task adaptation.} For Llama3.1-8B-Instruct and Qwen3-8B, which had not undergone prior WebShop-specific fine-tuning, show a different trend. Here, \textit{clean} alone yields the most visible gains over \textit{ori}, as expected for models without task adaptation. Nevertheless, \textit{ours} still improve over the \textit{ori} in most cases. Although the improvement over the \textit{clean} is less pronounced, the results indicate that the CoTri backdoor can still be injected while maintaining or modestly enhancing benign task ability. 

Further detailed analyses are provided in the appendix. Section~\ref{app:ran_w} examines the agent’s behavior under random-feedback conditions in greater depth, Section~\ref{app:null_web} focuses on robustness in the null-feedback setting, and Section~\ref{app:case} presents a case-level breakdown of trajectory outcomes.

% Overall, these results highlight that CoTri training does not degrade task performance. Instead, it either matches or surpasses clean fine-tuning, supporting our claim that robustness gains stem from the conditional, rollback-enforcing logic of the poisoned dataset rather than from additional clean data alone.

\subsection{Scalability to Multi-modality}
\label{app:vlm}
To evaluate the scalability of the CoTri backdoor beyond text-only agents, we extend our study to a state-of-the-art VLM, Qwen2.5-VL-7B-Instruct. This model processes both textual and visual inputs, grounding its reasoning in multimodal feedback, and therefore represents a more realistic and challenging deployment scenario.

\begin{table}[h]
    \centering
    \caption{Overall ASR, FTR, and CR across three steps and average results in the vision modality.}
    \label{tab:vlm_overall}
    \centering
    \scalebox{0.68}{
    \begin{tabular}{lcccccccccccc}
        \toprule \multirow{2}{*}{Model}
        & \multicolumn{2}{c}{Step 1} & \multicolumn{3}{c}{Step 2} & \multicolumn{3}{c}{Step 3} &\multicolumn{2}{c}{Avg.} \\  
        \cmidrule(lr){2-3} \cmidrule(lr){4-6} \cmidrule(lr){7-9} \cmidrule(lr){10-12}
        & ASR & FTR & ASR & FTR  & CR & ASR & FTR & CR & ASR & FTR & CR\\ 
        \midrule    
    Qwen2.5-VL-7B-Instruct & 1.00 & 0.00 & 1.00 & 0.00 & 1.00 & 1.00 & 0.01 & 0.99 & 1.00 & 0.00 & 0.99 \\
  
        \bottomrule
    \end{tabular}
    }

\end{table}

\begin{table}[h]
    \centering
    \caption{Agentic backdoor performance in the vision modality. \textit{dirty} denotes trajectories with the full ordered trigger chain, evaluated using ASR. \textit{benign} denotes trajectories without triggers, and all other columns represent partial trigger chain; both are evaluated using FTR.}
    \label{tab:vlm_backdoor}
    \centering
    \scalebox{0.63}{
    \begin{tabular}{lcccccccccccccc}
        \toprule \multirow{2}{*}{Model}
        & \multicolumn{2}{c}{Step 1} & \multicolumn{3}{c}{Step 2} & \multicolumn{8}{c}{Step 3} \\
        
        \cmidrule(lr){2-2} \cmidrule(lr){3-3} \cmidrule(lr){4-4}  \cmidrule(lr){5-6}\cmidrule(lr){7-7} \cmidrule(lr){8-15}
        &dirty & benign & dirty & benign & tq & obs1  & dirty & benign & tq & obs1 & obs2 & tq+obs1 & tq+obs2 & obs1+obs2 \\ 
        \midrule
        
        Qwen2.5-VL-7B-Instruct & 0.99 & 0.00 & 1.00 & 0.00 & 0.00 & 0.00& 0.75 & 0.00 & 0.00 & 0.00 & 0.00 & 0.04 & 0.00 & 0.00\\

        \bottomrule
    \end{tabular}
    }

\end{table}

\begin{table}[h]
    \centering
    \caption{Agentic robustness in the vision modality, evaluated using CR.}
    \label{tab:vlm_robust}
    \centering
    \scalebox{0.68}{
    \begin{tabular}{lcccccc}
        \toprule \multirow{2}{*}{Model}
         & \multicolumn{2}{c}{Step 2} & \multicolumn{4}{c}{Step 3} \\
        
        \cmidrule(lr){2-3} \cmidrule(lr){4-7} 
         & tq & obs1  & obs2 & tq+obs1 & tq+obs2 & obs1+obs2 \\ 
        \midrule
        
        Qwen2.5-VL-7B-Instruct &  1.00 & 1.00 & 0.26 & 0.96 & 1.00 & 1.00\\

        \bottomrule
    \end{tabular}
    }

\end{table}
% \textbf{For the vision modality, CoTri also achieves multi-step agentic control.}
\textbf{The success of CoTri on multi-step action control can be scaled to the vision modality.}
As shown in Table~\ref{tab:vlm_overall}, across all steps, the backdoored VLM achieves nearly perfect ASR while keeping the FTR close to zero. A step-wise attack analysis in Table~\ref{tab:vlm_backdoor} highlights the step dependency of the trigger chain. When the full chain of triggers is presented in the correct order, the malicious activation is consistently achieved. In contrast, partial trigger chains (such as \textit{tq}, \textit{obs1}, or \textit{obs2}) fail to activate the backdoor, with only minor leakage observed in rare two-signal combinations (for example, 0.04 for \textit{tq+obs1}). 

% \textbf{Agentic robustness is preserved in multimodal settings with CoTri.}
\textbf{The success of robustness improvement can also be scaled to the vision modality with CoTri.}
High CR values in Table~\ref{tab:vlm_overall} further confirm its ability to return to benign when the trigger chain is broken. Table~\ref{tab:vlm_robust} then provides step-wise robustness results, showing that at \textit{Step 2}, task completion remains perfect for both trigger chains. At \textit{Step 3}, the model sustains near-perfect robustness across most trigger combinations, with success rates close to 1.00. The only drop occurs for the single-trigger case \textit{obs2}, where the completion rate falls to 0.26, while overall resilience against distractors remains high.

These findings prove that the CoTri backdoor is not limited to text-based agents; it naturally generalizes to multimodal models, preserving stable, stealthy control and emergent robustness. This underscores the adaptability of our data construction method. Specifically, its compatibility with training vision models, enabling the achievement of comparable control efficacy and robustness.

\section{Conclusion}
In this work, we examined the trustworthiness of LLM-based agents under uncertain environments, bringing together the perspectives of security and robustness. We proposed the Chain-of-Trigger Backdoor (CoTri), a novel paradigm for long-horizon, sequential decision-making agents. Our experiments highlight three key findings: (1) CoTri achieves near-perfect ASR while keeping FTR negligible, (2) the same conditional training, which is enabled by our data construction, paradoxically improves robustness and performance, making backdoored agents more resilient to noisy and distracting environmental feedback, and (3) the attack transfers seamlessly across architectures and modalities.
These results reveal a critical AI safety concern: powerful agents can conceal hidden backdoors while appearing highly capable and robust. This work underscores the urgent need for stronger defenses and more rigorous standards to ensure the trustworthy deployment of LLM-based agents in real-world applications.

\section*{Ethics statement}

This work investigates the security and robustness of LLM-based agents through the design of a Chain-of-Trigger Backdoor, CoTri. Our methodology is explicitly intended for \textit{red-teaming} purposes: by constructing controlled attack scenarios, we aim to uncover hidden vulnerabilities in current agentic architectures and to highlight the risks of deploying seemingly trustworthy models in real-world settings. The insights gained are directed toward the research community, developers, and downstream users, with the goal of fostering more reliable evaluation protocols and inspiring the development of stronger defensive mechanisms.
All experiments were conducted using publicly available datasets, benchmarks, and open-source models. Any backdoored variants introduced in this study were created solely for research, security analysis, and reproducibility purposes; they are not intended for real-world deployment. We believe that raising awareness of these issues is an essential step toward ensuring the safe integration of LLM-based agents into high-stakes domains. Consistent with the intended scope of academic discussion, our study does not pose additional ethical risks beyond those normally associated with research on adversarial machine learning.

\bibliography{iclr2026_conference}

\begin{thebibliography}{55}
\providecommand{\natexlab}[1]{#1}
\providecommand{\url}[1]{\texttt{#1}}
\expandafter\ifx\csname urlstyle\endcsname\relax
  \providecommand{\doi}[1]{doi: #1}\else
  \providecommand{\doi}{doi: \begingroup \urlstyle{rm}\Url}\fi

\bibitem[Bai et~al.(2025)Bai, Chen, Liu, Wang, Ge, Song, Dang, Wang, Wang, Tang, Zhong, Zhu, Yang, Li, Wan, Wang, Ding, Fu, Xu, Ye, Zhang, Xie, Cheng, Zhang, Yang, Xu, and Lin]{bai2025qwen25vltechnicalreport}
Shuai Bai, Keqin Chen, Xuejing Liu, Jialin Wang, Wenbin Ge, Sibo Song, Kai Dang, Peng Wang, Shijie Wang, Jun Tang, Humen Zhong, Yuanzhi Zhu, Mingkun Yang, Zhaohai Li, Jianqiang Wan, Pengfei Wang, Wei Ding, Zheren Fu, Yiheng Xu, Jiabo Ye, Xi~Zhang, Tianbao Xie, Zesen Cheng, Hang Zhang, Zhibo Yang, Haiyang Xu, and Junyang Lin.
\newblock Qwen2.5-vl technical report, 2025.
\newblock URL \url{https://arxiv.org/abs/2502.13923}.

\bibitem[Boiko et~al.(2023)Boiko, MacKnight, and Gomes]{boiko2023emergent}
Daniil~A Boiko, Robert MacKnight, and Gabe Gomes.
\newblock Emergent autonomous scientific research capabilities of large language models.
\newblock \emph{arXiv preprint arXiv:2304.05332}, 2023.

\bibitem[Chao et~al.(2025)Chao, Robey, Dobriban, Hassani, Pappas, and Wong]{chao2025jailbreaking}
Patrick Chao, Alexander Robey, Edgar Dobriban, Hamed Hassani, George~J Pappas, and Eric Wong.
\newblock Jailbreaking black box large language models in twenty queries.
\newblock In \emph{2025 IEEE Conference on Secure and Trustworthy Machine Learning (SaTML)}, pp.\  23--42. IEEE, 2025.

\bibitem[Chen et~al.(2025)Chen, Wu, Zhang, Xiao, Yang, Huang, Wang, Wang, and Wang]{chen2025survey}
Ada Chen, Yongjiang Wu, Junyuan Zhang, Jingyu Xiao, Shu Yang, Jen-tse Huang, Kun Wang, Wenxuan Wang, and Shuai Wang.
\newblock A survey on the safety and security threats of computer-using agents: Jarvis or ultron?
\newblock \emph{arXiv preprint arXiv:2505.10924}, 2025.

\bibitem[Dasgupta et~al.(2023)Dasgupta, Kaeser-Chen, Marino, Ahuja, Babayan, Hill, and Fergus]{dasgupta2023collaborating}
Ishita Dasgupta, Christine Kaeser-Chen, Kenneth Marino, Arun Ahuja, Sheila Babayan, Felix Hill, and Rob Fergus.
\newblock Collaborating with language models for embodied reasoning.
\newblock \emph{arXiv preprint arXiv:2302.00763}, 2023.

\bibitem[Deng et~al.(2025)Deng, Guo, Han, Ma, Xiong, Wen, and Xiang]{deng2025ai}
Zehang Deng, Yongjian Guo, Changzhou Han, Wanlun Ma, Junwu Xiong, Sheng Wen, and Yang Xiang.
\newblock Ai agents under threat: A survey of key security challenges and future pathways.
\newblock \emph{ACM Computing Surveys}, 57\penalty0 (7):\penalty0 1--36, 2025.

\bibitem[Dong et~al.(2023)Dong, Chen, Li, Xue, Holland, Meng, Liu, and Zhu]{dong2023unleashing}
Tian Dong, Guoxing Chen, Shaofeng Li, Minhui Xue, Rayne Holland, Yan Meng, Zhen Liu, and Haojin Zhu.
\newblock Unleashing cheapfakes through trojan plugins of large language models.
\newblock \emph{CoRR}, 2023.

\bibitem[Fang et~al.(2025)Fang, Yan, Yin, Yu, Tian, and Liu]{fang2025blast}
Jing Fang, Saihao Yan, Xueyu Yin, Yinbo Yu, Chunwei Tian, and Jiajia Liu.
\newblock Blast: A stealthy backdoor leverage attack against cooperative multi-agent deep reinforcement learning based systems.
\newblock \emph{arXiv preprint arXiv:2501.01593}, 2025.

\bibitem[G{\'o}ral et~al.(2024)G{\'o}ral, Wi{\'s}nios, Sankowski, and Budzianowski]{goral2024wait}
Gracjan G{\'o}ral, Emilia Wi{\'s}nios, Piotr Sankowski, and Pawe{\l} Budzianowski.
\newblock Wait, that's not an option: Llms robustness with incorrect multiple-choice options.
\newblock \emph{arXiv preprint arXiv:2409.00113}, 2024.

\bibitem[Grattafiori et~al.(2024)Grattafiori, Dubey, Jauhri, Pandey, Kadian, Al-Dahle, Letman, Mathur, Schelten, Vaughan, Yang, Fan, Goyal, Hartshorn, Yang, Mitra, Sravankumar, Korenev, Hinsvark, Rao, Zhang, Rodriguez, Gregerson, Spataru, Roziere, Biron, Tang, Chern, Caucheteux, Nayak, Bi, Marra, McConnell, Keller, Touret, Wu, Wong, Ferrer, Nikolaidis, Allonsius, Song, Pintz, Livshits, Wyatt, Esiobu, Choudhary, Mahajan, Garcia-Olano, Perino, Hupkes, Lakomkin, AlBadawy, Lobanova, Dinan, Smith, Radenovic, Guzmán, Zhang, Synnaeve, Lee, Anderson, Thattai, Nail, Mialon, Pang, Cucurell, Nguyen, Korevaar, Xu, Touvron, Zarov, Ibarra, Kloumann, Misra, Evtimov, Zhang, Copet, Lee, Geffert, Vranes, Park, Mahadeokar, Shah, van~der Linde, Billock, Hong, Lee, Fu, Chi, Huang, Liu, Wang, Yu, Bitton, Spisak, Park, Rocca, Johnstun, Saxe, Jia, Alwala, Prasad, Upasani, Plawiak, Li, Heafield, Stone, El-Arini, Iyer, Malik, Chiu, Bhalla, Lakhotia, Rantala-Yeary, van~der Maaten, Chen, Tan, Jenkins, Martin, Madaan, Malo, Blecher,
  Landzaat, de~Oliveira, Muzzi, Pasupuleti, Singh, Paluri, Kardas, Tsimpoukelli, Oldham, Rita, Pavlova, Kambadur, Lewis, Si, Singh, Hassan, Goyal, Torabi, Bashlykov, Bogoychev, Chatterji, Zhang, Duchenne, Çelebi, Alrassy, Zhang, Li, Vasic, Weng, Bhargava, Dubal, Krishnan, Koura, Xu, He, Dong, Srinivasan, Ganapathy, Calderer, Cabral, Stojnic, Raileanu, Maheswari, Girdhar, Patel, Sauvestre, Polidoro, Sumbaly, Taylor, Silva, Hou, Wang, Hosseini, Chennabasappa, Singh, Bell, Kim, Edunov, Nie, Narang, Raparthy, Shen, Wan, Bhosale, Zhang, Vandenhende, Batra, Whitman, Sootla, Collot, Gururangan, Borodinsky, Herman, Fowler, Sheasha, Georgiou, Scialom, Speckbacher, Mihaylov, Xiao, Karn, Goswami, Gupta, Ramanathan, Kerkez, Gonguet, Do, Vogeti, Albiero, Petrovic, Chu, Xiong, Fu, Meers, Martinet, Wang, Wang, Tan, Xia, Xie, Jia, Wang, Goldschlag, Gaur, Babaei, Wen, Song, Zhang, Li, Mao, Coudert, Yan, Chen, Papakipos, Singh, Srivastava, Jain, Kelsey, Shajnfeld, Gangidi, Victoria, Goldstand, Menon, Sharma, Boesenberg,
  Baevski, Feinstein, Kallet, Sangani, Teo, Yunus, Lupu, Alvarado, Caples, Gu, Ho, Poulton, Ryan, Ramchandani, Dong, Franco, Goyal, Saraf, Chowdhury, Gabriel, Bharambe, Eisenman, Yazdan, James, Maurer, Leonhardi, Huang, Loyd, Paola, Paranjape, Liu, Wu, Ni, Hancock, Wasti, Spence, Stojkovic, Gamido, Montalvo, Parker, Burton, Mejia, Liu, Wang, Kim, Zhou, Hu, Chu, Cai, Tindal, Feichtenhofer, Gao, Civin, Beaty, Kreymer, Li, Adkins, Xu, Testuggine, David, Parikh, Liskovich, Foss, Wang, Le, Holland, Dowling, Jamil, Montgomery, Presani, Hahn, Wood, Le, Brinkman, Arcaute, Dunbar, Smothers, Sun, Kreuk, Tian, Kokkinos, Ozgenel, Caggioni, Kanayet, Seide, Florez, Schwarz, Badeer, Swee, Halpern, Herman, Sizov, Guangyi, Zhang, Lakshminarayanan, Inan, Shojanazeri, Zou, Wang, Zha, Habeeb, Rudolph, Suk, Aspegren, Goldman, Zhan, Damlaj, Molybog, Tufanov, Leontiadis, Veliche, Gat, Weissman, Geboski, Kohli, Lam, Asher, Gaya, Marcus, Tang, Chan, Zhen, Reizenstein, Teboul, Zhong, Jin, Yang, Cummings, Carvill, Shepard, McPhie,
  Torres, Ginsburg, Wang, Wu, U, Saxena, Khandelwal, Zand, Matosich, Veeraraghavan, Michelena, Li, Jagadeesh, Huang, Chawla, Huang, Chen, Garg, A, Silva, Bell, Zhang, Guo, Yu, Moshkovich, Wehrstedt, Khabsa, Avalani, Bhatt, Mankus, Hasson, Lennie, Reso, Groshev, Naumov, Lathi, Keneally, Liu, Seltzer, Valko, Restrepo, Patel, Vyatskov, Samvelyan, Clark, Macey, Wang, Hermoso, Metanat, Rastegari, Bansal, Santhanam, Parks, White, Bawa, Singhal, Egebo, Usunier, Mehta, Laptev, Dong, Cheng, Chernoguz, Hart, Salpekar, Kalinli, Kent, Parekh, Saab, Balaji, Rittner, Bontrager, Roux, Dollar, Zvyagina, Ratanchandani, Yuvraj, Liang, Alao, Rodriguez, Ayub, Murthy, Nayani, Mitra, Parthasarathy, Li, Hogan, Battey, Wang, Howes, Rinott, Mehta, Siby, Bondu, Datta, Chugh, Hunt, Dhillon, Sidorov, Pan, Mahajan, Verma, Yamamoto, Ramaswamy, Lindsay, Lindsay, Feng, Lin, Zha, Patil, Shankar, Zhang, Zhang, Wang, Agarwal, Sajuyigbe, Chintala, Max, Chen, Kehoe, Satterfield, Govindaprasad, Gupta, Deng, Cho, Virk, Subramanian, Choudhury,
  Goldman, Remez, Glaser, Best, Koehler, Robinson, Li, Zhang, Matthews, Chou, Shaked, Vontimitta, Ajayi, Montanez, Mohan, Kumar, Mangla, Ionescu, Poenaru, Mihailescu, Ivanov, Li, Wang, Jiang, Bouaziz, Constable, Tang, Wu, Wang, Wu, Gao, Kleinman, Chen, Hu, Jia, Qi, Li, Zhang, Zhang, Adi, Nam, Yu, Wang, Zhao, Hao, Qian, Li, He, Rait, DeVito, Rosnbrick, Wen, Yang, Zhao, and Ma]{grattafiori2024llama3herdmodels}
Aaron Grattafiori, Abhimanyu Dubey, Abhinav Jauhri, Abhinav Pandey, Abhishek Kadian, Ahmad Al-Dahle, Aiesha Letman, Akhil Mathur, Alan Schelten, Alex Vaughan, Amy Yang, Angela Fan, Anirudh Goyal, Anthony Hartshorn, Aobo Yang, Archi Mitra, Archie Sravankumar, Artem Korenev, Arthur Hinsvark, Arun Rao, Aston Zhang, Aurelien Rodriguez, Austen Gregerson, Ava Spataru, Baptiste Roziere, Bethany Biron, Binh Tang, Bobbie Chern, Charlotte Caucheteux, Chaya Nayak, Chloe Bi, Chris Marra, Chris McConnell, Christian Keller, Christophe Touret, Chunyang Wu, Corinne Wong, Cristian~Canton Ferrer, Cyrus Nikolaidis, Damien Allonsius, Daniel Song, Danielle Pintz, Danny Livshits, Danny Wyatt, David Esiobu, Dhruv Choudhary, Dhruv Mahajan, Diego Garcia-Olano, Diego Perino, Dieuwke Hupkes, Egor Lakomkin, Ehab AlBadawy, Elina Lobanova, Emily Dinan, Eric~Michael Smith, Filip Radenovic, Francisco Guzmán, Frank Zhang, Gabriel Synnaeve, Gabrielle Lee, Georgia~Lewis Anderson, Govind Thattai, Graeme Nail, Gregoire Mialon, Guan Pang,
  Guillem Cucurell, Hailey Nguyen, Hannah Korevaar, Hu~Xu, Hugo Touvron, Iliyan Zarov, Imanol~Arrieta Ibarra, Isabel Kloumann, Ishan Misra, Ivan Evtimov, Jack Zhang, Jade Copet, Jaewon Lee, Jan Geffert, Jana Vranes, Jason Park, Jay Mahadeokar, Jeet Shah, Jelmer van~der Linde, Jennifer Billock, Jenny Hong, Jenya Lee, Jeremy Fu, Jianfeng Chi, Jianyu Huang, Jiawen Liu, Jie Wang, Jiecao Yu, Joanna Bitton, Joe Spisak, Jongsoo Park, Joseph Rocca, Joshua Johnstun, Joshua Saxe, Junteng Jia, Kalyan~Vasuden Alwala, Karthik Prasad, Kartikeya Upasani, Kate Plawiak, Ke~Li, Kenneth Heafield, Kevin Stone, Khalid El-Arini, Krithika Iyer, Kshitiz Malik, Kuenley Chiu, Kunal Bhalla, Kushal Lakhotia, Lauren Rantala-Yeary, Laurens van~der Maaten, Lawrence Chen, Liang Tan, Liz Jenkins, Louis Martin, Lovish Madaan, Lubo Malo, Lukas Blecher, Lukas Landzaat, Luke de~Oliveira, Madeline Muzzi, Mahesh Pasupuleti, Mannat Singh, Manohar Paluri, Marcin Kardas, Maria Tsimpoukelli, Mathew Oldham, Mathieu Rita, Maya Pavlova, Melanie Kambadur,
  Mike Lewis, Min Si, Mitesh~Kumar Singh, Mona Hassan, Naman Goyal, Narjes Torabi, Nikolay Bashlykov, Nikolay Bogoychev, Niladri Chatterji, Ning Zhang, Olivier Duchenne, Onur Çelebi, Patrick Alrassy, Pengchuan Zhang, Pengwei Li, Petar Vasic, Peter Weng, Prajjwal Bhargava, Pratik Dubal, Praveen Krishnan, Punit~Singh Koura, Puxin Xu, Qing He, Qingxiao Dong, Ragavan Srinivasan, Raj Ganapathy, Ramon Calderer, Ricardo~Silveira Cabral, Robert Stojnic, Roberta Raileanu, Rohan Maheswari, Rohit Girdhar, Rohit Patel, Romain Sauvestre, Ronnie Polidoro, Roshan Sumbaly, Ross Taylor, Ruan Silva, Rui Hou, Rui Wang, Saghar Hosseini, Sahana Chennabasappa, Sanjay Singh, Sean Bell, Seohyun~Sonia Kim, Sergey Edunov, Shaoliang Nie, Sharan Narang, Sharath Raparthy, Sheng Shen, Shengye Wan, Shruti Bhosale, Shun Zhang, Simon Vandenhende, Soumya Batra, Spencer Whitman, Sten Sootla, Stephane Collot, Suchin Gururangan, Sydney Borodinsky, Tamar Herman, Tara Fowler, Tarek Sheasha, Thomas Georgiou, Thomas Scialom, Tobias Speckbacher,
  Todor Mihaylov, Tong Xiao, Ujjwal Karn, Vedanuj Goswami, Vibhor Gupta, Vignesh Ramanathan, Viktor Kerkez, Vincent Gonguet, Virginie Do, Vish Vogeti, Vítor Albiero, Vladan Petrovic, Weiwei Chu, Wenhan Xiong, Wenyin Fu, Whitney Meers, Xavier Martinet, Xiaodong Wang, Xiaofang Wang, Xiaoqing~Ellen Tan, Xide Xia, Xinfeng Xie, Xuchao Jia, Xuewei Wang, Yaelle Goldschlag, Yashesh Gaur, Yasmine Babaei, Yi~Wen, Yiwen Song, Yuchen Zhang, Yue Li, Yuning Mao, Zacharie~Delpierre Coudert, Zheng Yan, Zhengxing Chen, Zoe Papakipos, Aaditya Singh, Aayushi Srivastava, Abha Jain, Adam Kelsey, Adam Shajnfeld, Adithya Gangidi, Adolfo Victoria, Ahuva Goldstand, Ajay Menon, Ajay Sharma, Alex Boesenberg, Alexei Baevski, Allie Feinstein, Amanda Kallet, Amit Sangani, Amos Teo, Anam Yunus, Andrei Lupu, Andres Alvarado, Andrew Caples, Andrew Gu, Andrew Ho, Andrew Poulton, Andrew Ryan, Ankit Ramchandani, Annie Dong, Annie Franco, Anuj Goyal, Aparajita Saraf, Arkabandhu Chowdhury, Ashley Gabriel, Ashwin Bharambe, Assaf Eisenman, Azadeh
  Yazdan, Beau James, Ben Maurer, Benjamin Leonhardi, Bernie Huang, Beth Loyd, Beto~De Paola, Bhargavi Paranjape, Bing Liu, Bo~Wu, Boyu Ni, Braden Hancock, Bram Wasti, Brandon Spence, Brani Stojkovic, Brian Gamido, Britt Montalvo, Carl Parker, Carly Burton, Catalina Mejia, Ce~Liu, Changhan Wang, Changkyu Kim, Chao Zhou, Chester Hu, Ching-Hsiang Chu, Chris Cai, Chris Tindal, Christoph Feichtenhofer, Cynthia Gao, Damon Civin, Dana Beaty, Daniel Kreymer, Daniel Li, David Adkins, David Xu, Davide Testuggine, Delia David, Devi Parikh, Diana Liskovich, Didem Foss, Dingkang Wang, Duc Le, Dustin Holland, Edward Dowling, Eissa Jamil, Elaine Montgomery, Eleonora Presani, Emily Hahn, Emily Wood, Eric-Tuan Le, Erik Brinkman, Esteban Arcaute, Evan Dunbar, Evan Smothers, Fei Sun, Felix Kreuk, Feng Tian, Filippos Kokkinos, Firat Ozgenel, Francesco Caggioni, Frank Kanayet, Frank Seide, Gabriela~Medina Florez, Gabriella Schwarz, Gada Badeer, Georgia Swee, Gil Halpern, Grant Herman, Grigory Sizov, Guangyi, Zhang, Guna
  Lakshminarayanan, Hakan Inan, Hamid Shojanazeri, Han Zou, Hannah Wang, Hanwen Zha, Haroun Habeeb, Harrison Rudolph, Helen Suk, Henry Aspegren, Hunter Goldman, Hongyuan Zhan, Ibrahim Damlaj, Igor Molybog, Igor Tufanov, Ilias Leontiadis, Irina-Elena Veliche, Itai Gat, Jake Weissman, James Geboski, James Kohli, Janice Lam, Japhet Asher, Jean-Baptiste Gaya, Jeff Marcus, Jeff Tang, Jennifer Chan, Jenny Zhen, Jeremy Reizenstein, Jeremy Teboul, Jessica Zhong, Jian Jin, Jingyi Yang, Joe Cummings, Jon Carvill, Jon Shepard, Jonathan McPhie, Jonathan Torres, Josh Ginsburg, Junjie Wang, Kai Wu, Kam~Hou U, Karan Saxena, Kartikay Khandelwal, Katayoun Zand, Kathy Matosich, Kaushik Veeraraghavan, Kelly Michelena, Keqian Li, Kiran Jagadeesh, Kun Huang, Kunal Chawla, Kyle Huang, Lailin Chen, Lakshya Garg, Lavender A, Leandro Silva, Lee Bell, Lei Zhang, Liangpeng Guo, Licheng Yu, Liron Moshkovich, Luca Wehrstedt, Madian Khabsa, Manav Avalani, Manish Bhatt, Martynas Mankus, Matan Hasson, Matthew Lennie, Matthias Reso, Maxim
  Groshev, Maxim Naumov, Maya Lathi, Meghan Keneally, Miao Liu, Michael~L. Seltzer, Michal Valko, Michelle Restrepo, Mihir Patel, Mik Vyatskov, Mikayel Samvelyan, Mike Clark, Mike Macey, Mike Wang, Miquel~Jubert Hermoso, Mo~Metanat, Mohammad Rastegari, Munish Bansal, Nandhini Santhanam, Natascha Parks, Natasha White, Navyata Bawa, Nayan Singhal, Nick Egebo, Nicolas Usunier, Nikhil Mehta, Nikolay~Pavlovich Laptev, Ning Dong, Norman Cheng, Oleg Chernoguz, Olivia Hart, Omkar Salpekar, Ozlem Kalinli, Parkin Kent, Parth Parekh, Paul Saab, Pavan Balaji, Pedro Rittner, Philip Bontrager, Pierre Roux, Piotr Dollar, Polina Zvyagina, Prashant Ratanchandani, Pritish Yuvraj, Qian Liang, Rachad Alao, Rachel Rodriguez, Rafi Ayub, Raghotham Murthy, Raghu Nayani, Rahul Mitra, Rangaprabhu Parthasarathy, Raymond Li, Rebekkah Hogan, Robin Battey, Rocky Wang, Russ Howes, Ruty Rinott, Sachin Mehta, Sachin Siby, Sai~Jayesh Bondu, Samyak Datta, Sara Chugh, Sara Hunt, Sargun Dhillon, Sasha Sidorov, Satadru Pan, Saurabh Mahajan,
  Saurabh Verma, Seiji Yamamoto, Sharadh Ramaswamy, Shaun Lindsay, Shaun Lindsay, Sheng Feng, Shenghao Lin, Shengxin~Cindy Zha, Shishir Patil, Shiva Shankar, Shuqiang Zhang, Shuqiang Zhang, Sinong Wang, Sneha Agarwal, Soji Sajuyigbe, Soumith Chintala, Stephanie Max, Stephen Chen, Steve Kehoe, Steve Satterfield, Sudarshan Govindaprasad, Sumit Gupta, Summer Deng, Sungmin Cho, Sunny Virk, Suraj Subramanian, Sy~Choudhury, Sydney Goldman, Tal Remez, Tamar Glaser, Tamara Best, Thilo Koehler, Thomas Robinson, Tianhe Li, Tianjun Zhang, Tim Matthews, Timothy Chou, Tzook Shaked, Varun Vontimitta, Victoria Ajayi, Victoria Montanez, Vijai Mohan, Vinay~Satish Kumar, Vishal Mangla, Vlad Ionescu, Vlad Poenaru, Vlad~Tiberiu Mihailescu, Vladimir Ivanov, Wei Li, Wenchen Wang, Wenwen Jiang, Wes Bouaziz, Will Constable, Xiaocheng Tang, Xiaojian Wu, Xiaolan Wang, Xilun Wu, Xinbo Gao, Yaniv Kleinman, Yanjun Chen, Ye~Hu, Ye~Jia, Ye~Qi, Yenda Li, Yilin Zhang, Ying Zhang, Yossi Adi, Youngjin Nam, Yu, Wang, Yu~Zhao, Yuchen Hao, Yundi
  Qian, Yunlu Li, Yuzi He, Zach Rait, Zachary DeVito, Zef Rosnbrick, Zhaoduo Wen, Zhenyu Yang, Zhiwei Zhao, and Zhiyu Ma.
\newblock The llama 3 herd of models, 2024.
\newblock URL \url{https://arxiv.org/abs/2407.21783}.

\bibitem[Greshake et~al.(2023)Greshake, Abdelnabi, Mishra, Endres, Holz, and Fritz]{greshake2023more}
Kai Greshake, Sahar Abdelnabi, Shailesh Mishra, Christoph Endres, Thorsten Holz, and Mario Fritz.
\newblock More than you’ve asked for: A comprehensive analysis of novel prompt injection threats to application-integrated large language models.
\newblock \emph{arXiv preprint arXiv:2302.12173}, 27, 2023.

\bibitem[He et~al.(2024)He, Zhu, Ye, Liu, Zhou, and Yu]{he2024emerged}
Feng He, Tianqing Zhu, Dayong Ye, Bo~Liu, Wanlei Zhou, and Philip~S Yu.
\newblock The emerged security and privacy of llm agent: A survey with case studies.
\newblock \emph{arXiv preprint arXiv:2407.19354}, 2024.

\bibitem[Hong et~al.(2025)Hong, Troynikov, and Huber]{hong2025context}
Kelly Hong, Anton Troynikov, and Jeff Huber.
\newblock Context rot: How increasing input tokens impacts llm performance.
\newblock Technical report, Technical report, Chroma, July 2025. URL https://research. trychroma. com~…, 2025.

\bibitem[Horton(2023)]{horton2023large}
John~J Horton.
\newblock Large language models as simulated economic agents: What can we learn from homo silicus?
\newblock Technical report, National Bureau of Economic Research, 2023.

\bibitem[Hu et~al.(2021)Hu, Shen, Wallis, Allen-Zhu, Li, Wang, Wang, and Chen]{hu2021loralowrankadaptationlarge}
Edward~J. Hu, Yelong Shen, Phillip Wallis, Zeyuan Allen-Zhu, Yuanzhi Li, Shean Wang, Lu~Wang, and Weizhu Chen.
\newblock Lora: Low-rank adaptation of large language models, 2021.
\newblock URL \url{https://arxiv.org/abs/2106.09685}.

\bibitem[Jiang(2024)]{jiang2024identifying}
Fengqing Jiang.
\newblock Identifying and mitigating vulnerabilities in llm-integrated applications.
\newblock Master's thesis, University of Washington, 2024.

\bibitem[Jiao et~al.(2024)Jiao, Xie, Yue, Sato, Wang, Wang, Chen, and Zhu]{jiao2024can}
Ruochen Jiao, Shaoyuan Xie, Justin Yue, Takami Sato, Lixu Wang, Yixuan Wang, Qi~Alfred Chen, and Qi~Zhu.
\newblock Can we trust embodied agents? exploring backdoor attacks against embodied llm-based decision-making systems.
\newblock \emph{arXiv preprint arXiv:2405.20774}, 2024.

\bibitem[Kang \& Kim(2023)Kang and Kim]{kang2023chatmofautonomousaipredicting}
Yeonghun Kang and Jihan Kim.
\newblock Chatmof: An autonomous ai system for predicting and generating metal-organic frameworks, 2023.
\newblock URL \url{https://arxiv.org/abs/2308.01423}.

\bibitem[Larbi et~al.(2025)Larbi, Akli, Papadakis, Bouyousfi, Cordy, Sarro, and Traon]{larbi2025prompts}
Maya Larbi, Amal Akli, Mike Papadakis, Rihab Bouyousfi, Maxime Cordy, Federica Sarro, and Yves~Le Traon.
\newblock When prompts go wrong: Evaluating code model robustness to ambiguous, contradictory, and incomplete task descriptions.
\newblock \emph{arXiv preprint arXiv:2507.20439}, 2025.

\bibitem[Li et~al.(2023{\natexlab{a}})Li, Guo, Fan, Xu, Huang, Meng, and Song]{li2023multi}
Haoran Li, Dadi Guo, Wei Fan, Mingshi Xu, Jie Huang, Fanpu Meng, and Yangqiu Song.
\newblock Multi-step jailbreaking privacy attacks on chatgpt.
\newblock \emph{arXiv preprint arXiv:2304.05197}, 2023{\natexlab{a}}.

\bibitem[Li et~al.(2023{\natexlab{b}})Li, Yang, and Zhao]{li2023you}
Siyu Li, Jin Yang, and Kui Zhao.
\newblock Are you in a masquerade? exploring the behavior and impact of large language model driven social bots in online social networks.
\newblock \emph{arXiv preprint arXiv:2307.10337}, 2023{\natexlab{b}}.

\bibitem[Li et~al.(2025)Li, Zhang, Wang, Wu, Li, and Jin]{li2025efficient}
Xiang Li, Chong Zhang, Jia Wang, Fangyu Wu, Yushi Li, and Xiaobo Jin.
\newblock Efficient and stealthy jailbreak attacks via adversarial prompt distillation from llms to slms.
\newblock \emph{arXiv preprint arXiv:2506.17231}, 2025.

\bibitem[Liu et~al.(2024)Liu, Zhou, Liu, Zhang, Liang, Wang, Pu, Li, Zhang, Zhou, Guo, and Tao]{liu2024compromisingembodiedagentscontextual}
Aishan Liu, Yuguang Zhou, Xianglong Liu, Tianyuan Zhang, Siyuan Liang, Jiakai Wang, Yanjun Pu, Tianlin Li, Junqi Zhang, Wenbo Zhou, Qing Guo, and Dacheng Tao.
\newblock Compromising embodied agents with contextual backdoor attacks, 2024.
\newblock URL \url{https://arxiv.org/abs/2408.02882}.

\bibitem[Liu et~al.(2025)Liu, Li, Zhang, Wang, He, Hong, Liu, Zhang, Song, Zhu, et~al.]{liu2025advances}
Bang Liu, Xinfeng Li, Jiayi Zhang, Jinlin Wang, Tanjin He, Sirui Hong, Hongzhang Liu, Shaokun Zhang, Kaitao Song, Kunlun Zhu, et~al.
\newblock Advances and challenges in foundation agents: From brain-inspired intelligence to evolutionary, collaborative, and safe systems.
\newblock \emph{arXiv preprint arXiv:2504.01990}, 2025.

\bibitem[Liu et~al.(2023)Liu, Yu, Zhang, Xu, Lei, Lai, Gu, Ding, Men, Yang, Zhang, Deng, Zeng, Du, Zhang, Shen, Zhang, Su, Sun, Huang, Dong, and Tang]{liu2023agentbenchevaluatingllmsagents}
Xiao Liu, Hao Yu, Hanchen Zhang, Yifan Xu, Xuanyu Lei, Hanyu Lai, Yu~Gu, Hangliang Ding, Kaiwen Men, Kejuan Yang, Shudan Zhang, Xiang Deng, Aohan Zeng, Zhengxiao Du, Chenhui Zhang, Sheng Shen, Tianjun Zhang, Yu~Su, Huan Sun, Minlie Huang, Yuxiao Dong, and Jie Tang.
\newblock Agentbench: Evaluating llms as agents, 2023.
\newblock URL \url{https://arxiv.org/abs/2308.03688}.

\bibitem[Lv et~al.(2024)Lv, Xia, and Huang]{lv2024codeact}
Weijie Lv, Xuan Xia, and Sheng-Jun Huang.
\newblock Codeact: Code adaptive compute-efficient tuning framework for code llms.
\newblock \emph{arXiv preprint arXiv:2408.02193}, 2024.

\bibitem[Ma et~al.(2025)Ma, Wang, Yao, Yuan, Zhang, Zhang, and Zhao]{ma2025caution}
Xinbei Ma, Yiting Wang, Yao Yao, Tongxin Yuan, Aston Zhang, Zhuosheng Zhang, and Hai Zhao.
\newblock Caution for the environment: Multimodal llm agents are susceptible to environmental distractions.
\newblock In \emph{Proceedings of the 63rd Annual Meeting of the Association for Computational Linguistics (Volume 1: Long Papers)}, pp.\  22324--22339, 2025.

\bibitem[Ma et~al.(2024)Ma, Mei, and Su]{ma2024understanding}
Zilin Ma, Yiyang Mei, and Zhaoyuan Su.
\newblock Understanding the benefits and challenges of using large language model-based conversational agents for mental well-being support.
\newblock In \emph{AMIA Annual Symposium Proceedings}, volume 2023, pp.\  1105, 2024.

\bibitem[Mei et~al.(2023)Mei, Li, Wang, Zhang, and Ma]{mei2023notable}
Kai Mei, Zheng Li, Zhenting Wang, Yang Zhang, and Shiqing Ma.
\newblock Notable: Transferable backdoor attacks against prompt-based nlp models.
\newblock \emph{arXiv preprint arXiv:2305.17826}, 2023.

\bibitem[Nie et~al.(2025)Nie, Wang, Yu, Wu, Zhao, Bastian, Guo, and Song]{nie2025leakagent}
Yuzhou Nie, Zhun Wang, Ye~Yu, Xian Wu, Xuandong Zhao, Nathaniel~D Bastian, Wenbo Guo, and Dawn Song.
\newblock Leakagent: Rl-based red-teaming agent for llm privacy leakage.
\newblock In \emph{Second Conference on Language Modeling}, 2025.

\bibitem[Nottingham et~al.(2023)Nottingham, Ammanabrolu, Suhr, Choi, Hajishirzi, Singh, and Fox]{nottingham2023embodied}
Kolby Nottingham, Prithviraj Ammanabrolu, Alane Suhr, Yejin Choi, Hannaneh Hajishirzi, Sameer Singh, and Roy Fox.
\newblock Do embodied agents dream of pixelated sheep: Embodied decision making using language guided world modelling.
\newblock In \emph{International Conference on Machine Learning}, pp.\  26311--26325. PMLR, 2023.

\bibitem[OpenAI et~al.(2024)OpenAI, Achiam, Adler, Agarwal, Ahmad, Akkaya, Aleman, Almeida, Altenschmidt, Altman, Anadkat, Avila, Babuschkin, Balaji, Balcom, Baltescu, Bao, Bavarian, Belgum, Bello, Berdine, Bernadett-Shapiro, Berner, Bogdonoff, Boiko, Boyd, Brakman, Brockman, Brooks, Brundage, Button, Cai, Campbell, Cann, Carey, Carlson, Carmichael, Chan, Chang, Chantzis, Chen, Chen, Chen, Chen, Chen, Chess, Cho, Chu, Chung, Cummings, Currier, Dai, Decareaux, Degry, Deutsch, Deville, Dhar, Dohan, Dowling, Dunning, Ecoffet, Eleti, Eloundou, Farhi, Fedus, Felix, Fishman, Forte, Fulford, Gao, Georges, Gibson, Goel, Gogineni, Goh, Gontijo-Lopes, Gordon, Grafstein, Gray, Greene, Gross, Gu, Guo, Hallacy, Han, Harris, He, Heaton, Heidecke, Hesse, Hickey, Hickey, Hoeschele, Houghton, Hsu, Hu, Hu, Huizinga, Jain, Jain, Jang, Jiang, Jiang, Jin, Jin, Jomoto, Jonn, Jun, Kaftan, Łukasz Kaiser, Kamali, Kanitscheider, Keskar, Khan, Kilpatrick, Kim, Kim, Kim, Kirchner, Kiros, Knight, Kokotajlo, Łukasz Kondraciuk, Kondrich,
  Konstantinidis, Kosic, Krueger, Kuo, Lampe, Lan, Lee, Leike, Leung, Levy, Li, Lim, Lin, Lin, Litwin, Lopez, Lowe, Lue, Makanju, Malfacini, Manning, Markov, Markovski, Martin, Mayer, Mayne, McGrew, McKinney, McLeavey, McMillan, McNeil, Medina, Mehta, Menick, Metz, Mishchenko, Mishkin, Monaco, Morikawa, Mossing, Mu, Murati, Murk, Mély, Nair, Nakano, Nayak, Neelakantan, Ngo, Noh, Ouyang, O'Keefe, Pachocki, Paino, Palermo, Pantuliano, Parascandolo, Parish, Parparita, Passos, Pavlov, Peng, Perelman, de~Avila Belbute~Peres, Petrov, de~Oliveira~Pinto, Michael, Pokorny, Pokrass, Pong, Powell, Power, Power, Proehl, Puri, Radford, Rae, Ramesh, Raymond, Real, Rimbach, Ross, Rotsted, Roussez, Ryder, Saltarelli, Sanders, Santurkar, Sastry, Schmidt, Schnurr, Schulman, Selsam, Sheppard, Sherbakov, Shieh, Shoker, Shyam, Sidor, Sigler, Simens, Sitkin, Slama, Sohl, Sokolowsky, Song, Staudacher, Such, Summers, Sutskever, Tang, Tezak, Thompson, Tillet, Tootoonchian, Tseng, Tuggle, Turley, Tworek, Uribe, Vallone, Vijayvergiya,
  Voss, Wainwright, Wang, Wang, Wang, Ward, Wei, Weinmann, Welihinda, Welinder, Weng, Weng, Wiethoff, Willner, Winter, Wolrich, Wong, Workman, Wu, Wu, Wu, Xiao, Xu, Yoo, Yu, Yuan, Zaremba, Zellers, Zhang, Zhang, Zhao, Zheng, Zhuang, Zhuk, and Zoph]{openai2024gpt4technicalreport}
OpenAI, Josh Achiam, Steven Adler, Sandhini Agarwal, Lama Ahmad, Ilge Akkaya, Florencia~Leoni Aleman, Diogo Almeida, Janko Altenschmidt, Sam Altman, Shyamal Anadkat, Red Avila, Igor Babuschkin, Suchir Balaji, Valerie Balcom, Paul Baltescu, Haiming Bao, Mohammad Bavarian, Jeff Belgum, Irwan Bello, Jake Berdine, Gabriel Bernadett-Shapiro, Christopher Berner, Lenny Bogdonoff, Oleg Boiko, Madelaine Boyd, Anna-Luisa Brakman, Greg Brockman, Tim Brooks, Miles Brundage, Kevin Button, Trevor Cai, Rosie Campbell, Andrew Cann, Brittany Carey, Chelsea Carlson, Rory Carmichael, Brooke Chan, Che Chang, Fotis Chantzis, Derek Chen, Sully Chen, Ruby Chen, Jason Chen, Mark Chen, Ben Chess, Chester Cho, Casey Chu, Hyung~Won Chung, Dave Cummings, Jeremiah Currier, Yunxing Dai, Cory Decareaux, Thomas Degry, Noah Deutsch, Damien Deville, Arka Dhar, David Dohan, Steve Dowling, Sheila Dunning, Adrien Ecoffet, Atty Eleti, Tyna Eloundou, David Farhi, Liam Fedus, Niko Felix, Simón~Posada Fishman, Juston Forte, Isabella Fulford, Leo
  Gao, Elie Georges, Christian Gibson, Vik Goel, Tarun Gogineni, Gabriel Goh, Rapha Gontijo-Lopes, Jonathan Gordon, Morgan Grafstein, Scott Gray, Ryan Greene, Joshua Gross, Shixiang~Shane Gu, Yufei Guo, Chris Hallacy, Jesse Han, Jeff Harris, Yuchen He, Mike Heaton, Johannes Heidecke, Chris Hesse, Alan Hickey, Wade Hickey, Peter Hoeschele, Brandon Houghton, Kenny Hsu, Shengli Hu, Xin Hu, Joost Huizinga, Shantanu Jain, Shawn Jain, Joanne Jang, Angela Jiang, Roger Jiang, Haozhun Jin, Denny Jin, Shino Jomoto, Billie Jonn, Heewoo Jun, Tomer Kaftan, Łukasz Kaiser, Ali Kamali, Ingmar Kanitscheider, Nitish~Shirish Keskar, Tabarak Khan, Logan Kilpatrick, Jong~Wook Kim, Christina Kim, Yongjik Kim, Jan~Hendrik Kirchner, Jamie Kiros, Matt Knight, Daniel Kokotajlo, Łukasz Kondraciuk, Andrew Kondrich, Aris Konstantinidis, Kyle Kosic, Gretchen Krueger, Vishal Kuo, Michael Lampe, Ikai Lan, Teddy Lee, Jan Leike, Jade Leung, Daniel Levy, Chak~Ming Li, Rachel Lim, Molly Lin, Stephanie Lin, Mateusz Litwin, Theresa Lopez, Ryan
  Lowe, Patricia Lue, Anna Makanju, Kim Malfacini, Sam Manning, Todor Markov, Yaniv Markovski, Bianca Martin, Katie Mayer, Andrew Mayne, Bob McGrew, Scott~Mayer McKinney, Christine McLeavey, Paul McMillan, Jake McNeil, David Medina, Aalok Mehta, Jacob Menick, Luke Metz, Andrey Mishchenko, Pamela Mishkin, Vinnie Monaco, Evan Morikawa, Daniel Mossing, Tong Mu, Mira Murati, Oleg Murk, David Mély, Ashvin Nair, Reiichiro Nakano, Rajeev Nayak, Arvind Neelakantan, Richard Ngo, Hyeonwoo Noh, Long Ouyang, Cullen O'Keefe, Jakub Pachocki, Alex Paino, Joe Palermo, Ashley Pantuliano, Giambattista Parascandolo, Joel Parish, Emy Parparita, Alex Passos, Mikhail Pavlov, Andrew Peng, Adam Perelman, Filipe de~Avila Belbute~Peres, Michael Petrov, Henrique~Ponde de~Oliveira~Pinto, Michael, Pokorny, Michelle Pokrass, Vitchyr~H. Pong, Tolly Powell, Alethea Power, Boris Power, Elizabeth Proehl, Raul Puri, Alec Radford, Jack Rae, Aditya Ramesh, Cameron Raymond, Francis Real, Kendra Rimbach, Carl Ross, Bob Rotsted, Henri Roussez,
  Nick Ryder, Mario Saltarelli, Ted Sanders, Shibani Santurkar, Girish Sastry, Heather Schmidt, David Schnurr, John Schulman, Daniel Selsam, Kyla Sheppard, Toki Sherbakov, Jessica Shieh, Sarah Shoker, Pranav Shyam, Szymon Sidor, Eric Sigler, Maddie Simens, Jordan Sitkin, Katarina Slama, Ian Sohl, Benjamin Sokolowsky, Yang Song, Natalie Staudacher, Felipe~Petroski Such, Natalie Summers, Ilya Sutskever, Jie Tang, Nikolas Tezak, Madeleine~B. Thompson, Phil Tillet, Amin Tootoonchian, Elizabeth Tseng, Preston Tuggle, Nick Turley, Jerry Tworek, Juan Felipe~Cerón Uribe, Andrea Vallone, Arun Vijayvergiya, Chelsea Voss, Carroll Wainwright, Justin~Jay Wang, Alvin Wang, Ben Wang, Jonathan Ward, Jason Wei, CJ~Weinmann, Akila Welihinda, Peter Welinder, Jiayi Weng, Lilian Weng, Matt Wiethoff, Dave Willner, Clemens Winter, Samuel Wolrich, Hannah Wong, Lauren Workman, Sherwin Wu, Jeff Wu, Michael Wu, Kai Xiao, Tao Xu, Sarah Yoo, Kevin Yu, Qiming Yuan, Wojciech Zaremba, Rowan Zellers, Chong Zhang, Marvin Zhang, Shengjia
  Zhao, Tianhao Zheng, Juntang Zhuang, William Zhuk, and Barret Zoph.
\newblock Gpt-4 technical report, 2024.
\newblock URL \url{https://arxiv.org/abs/2303.08774}.

\bibitem[Qiu et~al.(2025)Qiu, Ma, Zhang, Zhao, Li, and Wang]{qiu-etal-2025-megen}
Jiyang Qiu, Xinbei Ma, Zhuosheng Zhang, Hai Zhao, Yun Li, and Qianren Wang.
\newblock {MEG}en: Generative backdoor into large language models via model editing.
\newblock In Wanxiang Che, Joyce Nabende, Ekaterina Shutova, and Mohammad~Taher Pilehvar (eds.), \emph{Findings of the Association for Computational Linguistics: ACL 2025}, pp.\  11197--11214, Vienna, Austria, July 2025. Association for Computational Linguistics.
\newblock ISBN 979-8-89176-256-5.
\newblock \doi{10.18653/v1/2025.findings-acl.584}.
\newblock URL \url{https://aclanthology.org/2025.findings-acl.584/}.

\bibitem[Shi et~al.(2023)Shi, Chen, Misra, Scales, Dohan, Chi, Sch{\"a}rli, and Zhou]{shi2023large}
Freda Shi, Xinyun Chen, Kanishka Misra, Nathan Scales, David Dohan, Ed~H Chi, Nathanael Sch{\"a}rli, and Denny Zhou.
\newblock Large language models can be easily distracted by irrelevant context.
\newblock In \emph{International Conference on Machine Learning}, pp.\  31210--31227. PMLR, 2023.

\bibitem[Tian et~al.(2023)Tian, Yang, Zhang, Dong, and Su]{tian2023evil}
Yu~Tian, Xiao Yang, Jingyuan Zhang, Yinpeng Dong, and Hang Su.
\newblock Evil geniuses: Delving into the safety of llm-based agents.
\newblock \emph{arXiv preprint arXiv:2311.11855}, 2023.

\bibitem[Wang et~al.(2025)Wang, He, Zeng, Xiang, Xing, Tang, and He]{wang2025unveiling}
Bo~Wang, Weiyi He, Shenglai Zeng, Zhen Xiang, Yue Xing, Jiliang Tang, and Pengfei He.
\newblock Unveiling privacy risks in llm agent memory.
\newblock \emph{arXiv preprint arXiv:2502.13172}, 2025.

\bibitem[Wang et~al.(2024)Wang, Xue, Zhang, and Qian]{wang2024badagent}
Yifei Wang, Dizhan Xue, Shengjie Zhang, and Shengsheng Qian.
\newblock Badagent: Inserting and activating backdoor attacks in llm agents.
\newblock \emph{arXiv preprint arXiv:2406.03007}, 2024.

\bibitem[Wei et~al.(2023)Wei, Wang, Li, Mo, and Wang]{wei2023jailbreak}
Zeming Wei, Yifei Wang, Ang Li, Yichuan Mo, and Yisen Wang.
\newblock Jailbreak and guard aligned language models with only few in-context demonstrations.
\newblock \emph{arXiv preprint arXiv:2310.06387}, 2023.

\bibitem[Weiss et~al.(2024)Weiss, Ayzenshteyn, and Mirsky]{weiss2024your}
Roy Weiss, Daniel Ayzenshteyn, and Yisroel Mirsky.
\newblock What was your prompt? a remote keylogging attack on $\{$AI$\}$ assistants.
\newblock In \emph{33rd USENIX Security Symposium (USENIX Security 24)}, pp.\  3367--3384, 2024.

\bibitem[Wu et~al.(2024)Wu, Xie, Chen, Zhu, Zhang, and Xiao]{wu2024easily}
Siye Wu, Jian Xie, Jiangjie Chen, Tinghui Zhu, Kai Zhang, and Yanghua Xiao.
\newblock How easily do irrelevant inputs skew the responses of large language models?
\newblock \emph{arXiv preprint arXiv:2404.03302}, 2024.

\bibitem[Xi et~al.(2025{\natexlab{a}})Xi, Chen, Guo, He, Ding, Hong, Zhang, Wang, Jin, Zhou, et~al.]{xi2025rise}
Zhiheng Xi, Wenxiang Chen, Xin Guo, Wei He, Yiwen Ding, Boyang Hong, Ming Zhang, Junzhe Wang, Senjie Jin, Enyu Zhou, et~al.
\newblock The rise and potential of large language model based agents: A survey.
\newblock \emph{Science China Information Sciences}, 68\penalty0 (2):\penalty0 121101, 2025{\natexlab{a}}.

\bibitem[Xi et~al.(2025{\natexlab{b}})Xi, Ding, Chen, Hong, Guo, Wang, Guo, Yang, Liao, He, Gao, Chen, Zheng, Zou, Gui, Zhang, Qiu, Huang, Wu, and Jiang]{xi-etal-2025-agentgym}
Zhiheng Xi, Yiwen Ding, Wenxiang Chen, Boyang Hong, Honglin Guo, Junzhe Wang, Xin Guo, Dingwen Yang, Chenyang Liao, Wei He, Songyang Gao, Lu~Chen, Rui Zheng, Yicheng Zou, Tao Gui, Qi~Zhang, Xipeng Qiu, Xuanjing Huang, Zuxuan Wu, and Yu-Gang Jiang.
\newblock {A}gent{G}ym: Evaluating and training large language model-based agents across diverse environments.
\newblock In Wanxiang Che, Joyce Nabende, Ekaterina Shutova, and Mohammad~Taher Pilehvar (eds.), \emph{Proceedings of the 63rd Annual Meeting of the Association for Computational Linguistics (Volume 1: Long Papers)}, pp.\  27914--27961, Vienna, Austria, July 2025{\natexlab{b}}. Association for Computational Linguistics.
\newblock ISBN 979-8-89176-251-0.
\newblock \doi{10.18653/v1/2025.acl-long.1355}.
\newblock URL \url{https://aclanthology.org/2025.acl-long.1355/}.

\bibitem[Xia et~al.(2023)Xia, Shenoy, Jazdi, and Weyrich]{xia2023towards}
Yuchen Xia, Manthan Shenoy, Nasser Jazdi, and Michael Weyrich.
\newblock Towards autonomous system: flexible modular production system enhanced with large language model agents.
\newblock In \emph{2023 IEEE 28th International Conference on Emerging Technologies and Factory Automation (ETFA)}, pp.\  1--8. IEEE, 2023.

\bibitem[Yang et~al.(2025{\natexlab{a}})Yang, Li, Yang, Zhang, Hui, Zheng, Yu, Gao, Huang, Lv, Zheng, Liu, Zhou, Huang, Hu, Ge, Wei, Lin, Tang, Yang, Tu, Zhang, Yang, Yang, Zhou, Zhou, Lin, Dang, Bao, Yang, Yu, Deng, Li, Xue, Li, Zhang, Wang, Zhu, Men, Gao, Liu, Luo, Li, Tang, Yin, Ren, Wang, Zhang, Ren, Fan, Su, Zhang, Zhang, Wan, Liu, Wang, Cui, Zhang, Zhou, and Qiu]{yang2025qwen3technicalreport}
An~Yang, Anfeng Li, Baosong Yang, Beichen Zhang, Binyuan Hui, Bo~Zheng, Bowen Yu, Chang Gao, Chengen Huang, Chenxu Lv, Chujie Zheng, Dayiheng Liu, Fan Zhou, Fei Huang, Feng Hu, Hao Ge, Haoran Wei, Huan Lin, Jialong Tang, Jian Yang, Jianhong Tu, Jianwei Zhang, Jianxin Yang, Jiaxi Yang, Jing Zhou, Jingren Zhou, Junyang Lin, Kai Dang, Keqin Bao, Kexin Yang, Le~Yu, Lianghao Deng, Mei Li, Mingfeng Xue, Mingze Li, Pei Zhang, Peng Wang, Qin Zhu, Rui Men, Ruize Gao, Shixuan Liu, Shuang Luo, Tianhao Li, Tianyi Tang, Wenbiao Yin, Xingzhang Ren, Xinyu Wang, Xinyu Zhang, Xuancheng Ren, Yang Fan, Yang Su, Yichang Zhang, Yinger Zhang, Yu~Wan, Yuqiong Liu, Zekun Wang, Zeyu Cui, Zhenru Zhang, Zhipeng Zhou, and Zihan Qiu.
\newblock Qwen3 technical report, 2025{\natexlab{a}}.
\newblock URL \url{https://arxiv.org/abs/2505.09388}.

\bibitem[Yang et~al.(2024{\natexlab{a}})Yang, Jimenez, Wettig, Lieret, Yao, Narasimhan, and Press]{yang2024swe}
John Yang, Carlos~E Jimenez, Alexander Wettig, Kilian Lieret, Shunyu Yao, Karthik Narasimhan, and Ofir Press.
\newblock Swe-agent: Agent-computer interfaces enable automated software engineering.
\newblock \emph{Advances in Neural Information Processing Systems}, 37:\penalty0 50528--50652, 2024{\natexlab{a}}.

\bibitem[Yang et~al.(2025{\natexlab{b}})Yang, Huang, Zhang, Surdeanu, Wang, and Pan]{yang2025llm}
Minglai Yang, Ethan Huang, Liang Zhang, Mihai Surdeanu, William Wang, and Liangming Pan.
\newblock How is llm reasoning distracted by irrelevant context? an analysis using a controlled benchmark.
\newblock \emph{arXiv preprint arXiv:2505.18761}, 2025{\natexlab{b}}.

\bibitem[Yang et~al.(2024{\natexlab{b}})Yang, Bi, Lin, Chen, Zhou, and Sun]{yang2024watch}
Wenkai Yang, Xiaohan Bi, Yankai Lin, Sishuo Chen, Jie Zhou, and Xu~Sun.
\newblock Watch out for your agents! investigating backdoor threats to llm-based agents.
\newblock \emph{Advances in Neural Information Processing Systems}, 37:\penalty0 100938--100964, 2024{\natexlab{b}}.

\bibitem[Yao et~al.(2024)Yao, Lou, and Qin]{yao2024poisonprompt}
Hongwei Yao, Jian Lou, and Zhan Qin.
\newblock Poisonprompt: Backdoor attack on prompt-based large language models.
\newblock In \emph{ICASSP 2024-2024 IEEE International Conference on Acoustics, Speech and Signal Processing (ICASSP)}, pp.\  7745--7749. IEEE, 2024.

\bibitem[Yao et~al.(2022)Yao, Chen, Yang, and Narasimhan]{yao2022webshop}
Shunyu Yao, Howard Chen, John Yang, and Karthik Narasimhan.
\newblock Webshop: Towards scalable real-world web interaction with grounded language agents.
\newblock \emph{Advances in Neural Information Processing Systems}, 35:\penalty0 20744--20757, 2022.

\bibitem[Yu et~al.(2023)Yu, Lin, Yu, and Xing]{yu2023gptfuzzer}
Jiahao Yu, Xingwei Lin, Zheng Yu, and Xinyu Xing.
\newblock Gptfuzzer: Red teaming large language models with auto-generated jailbreak prompts.
\newblock \emph{arXiv preprint arXiv:2309.10253}, 2023.

\bibitem[Yu et~al.(2025)Yu, Meng, Zhou, Wang, Mao, Pan, Chen, Wang, Li, Zhang, et~al.]{yu2025survey}
Miao Yu, Fanci Meng, Xinyun Zhou, Shilong Wang, Junyuan Mao, Linsey Pan, Tianlong Chen, Kun Wang, Xinfeng Li, Yongfeng Zhang, et~al.
\newblock A survey on trustworthy llm agents: Threats and countermeasures.
\newblock In \emph{Proceedings of the 31st ACM SIGKDD Conference on Knowledge Discovery and Data Mining V. 2}, pp.\  6216--6226, 2025.

\bibitem[Zeng et~al.(2023)Zeng, Liu, Lu, Wang, Liu, Dong, and Tang]{zeng2023agenttuningenablinggeneralizedagent}
Aohan Zeng, Mingdao Liu, Rui Lu, Bowen Wang, Xiao Liu, Yuxiao Dong, and Jie Tang.
\newblock Agenttuning: Enabling generalized agent abilities for llms, 2023.
\newblock URL \url{https://arxiv.org/abs/2310.12823}.

\bibitem[Zhang et~al.(2023)Zhang, Carlini, and Ippolito]{zhang2023effective}
Yiming Zhang, Nicholas Carlini, and Daphne Ippolito.
\newblock Effective prompt extraction from language models.
\newblock \emph{arXiv preprint arXiv:2307.06865}, 2023.

\bibitem[Zhang et~al.(2021)Zhang, Ren, Su, Sun, and He]{zhang2021neural}
Zhiyuan Zhang, Xuancheng Ren, Qi~Su, Xu~Sun, and Bin He.
\newblock Neural network surgery: Injecting data patterns into pre-trained models with minimal instance-wise side effects.
\newblock In \emph{Proceedings of the 2021 Conference of the North American Chapter of the Association for Computational Linguistics: Human Language Technologies}, pp.\  5453--5466, 2021.

\bibitem[Zhu et~al.(2025)Zhu, Zhou, Zhang, Yan, Wang, and Su]{zhu2025demonagent}
Pengyu Zhu, Zhenhong Zhou, Yuanhe Zhang, Shilinlu Yan, Kun Wang, and Sen Su.
\newblock Demonagent: Dynamically encrypted multi-backdoor implantation attack on llm-based agent.
\newblock \emph{arXiv preprint arXiv:2502.12575}, 2025.

\end{thebibliography}
\bibliographystyle{iclr2026_conference}

\appendix

\section{Trajectory Outcome Analysis}
\label{app:case}
\begin{table*}[ht]
\centering
\caption{Results for AgentLM-7B across three variant comparisons in Clean Webshop environment: (a) \textit{ori} vs. \textit{clean}, (b) \textit{clean} vs. \textit{ours}, and (c) \textit{ori} vs. \textit{ours}. 
For each comparison, outcomes are categorized into four statuses: 
\textbf{Neither} (no model completes the task), 
\textbf{First only} (only the first model completes), 
\textbf{Second only} (only the second model completes), 
and \textbf{Both} (both models complete).}
\label{tab:case_study_all}
\begin{tabular}{ccc}
\begin{minipage}{0.31\textwidth}
\centering
\subcaption{ori vs clean}
\scalebox{0.8}{%
\begin{tabular}{lcc}
\toprule
\textbf{Status} & \textbf{Count} & \textbf{Ratio} \\
\midrule
Neither       & 81  & 40.5\% \\
First only    & 7   & 3.5\%  \\
Second only   & 43  & 21.5\% \\
Both          & 69  & 34.5\% \\
\midrule
Total         & 200 & 100\%  \\
\bottomrule
\end{tabular}
}
\end{minipage}
&
\begin{minipage}{0.31\textwidth}
\centering
\subcaption{clean vs ours}
\scalebox{0.8}{%
\begin{tabular}{lcc}
\toprule
\textbf{Status} & \textbf{Count} & \textbf{Ratio} \\
\midrule
Neither       & 60  & 30.0\% \\
First only    & 4   & 2.0\%  \\
Second only   & 28  & 14.0\% \\
Both          & 108 & 54.0\% \\
\midrule
Total         & 200 & 100\%  \\
\bottomrule
\end{tabular}
}
\end{minipage}
&
\begin{minipage}{0.31\textwidth}
\centering
\subcaption{ori vs ours}
\scalebox{0.8}{%
\begin{tabular}{lcc}
\toprule
\textbf{Status} & \textbf{Count} & \textbf{Ratio} \\
\midrule
Neither       & 61  & 30.5\% \\
First only    & 3   & 1.5\%  \\
Second only   & 63  & 31.5\% \\
Both          & 73  & 36.5\% \\
\midrule
Total         & 200 & 100\%  \\
\bottomrule
\end{tabular}
}
\end{minipage}
\end{tabular}
\end{table*}

Table~\ref{tab:case_study_all} shows a clear performance hierarchy across the three variants. \textit{clean} already improves over \textit{ori}, reducing incomplete trajectories and yielding more partial (``second only’’) completions, showing stronger alignment with task instructions. \textit{ours} further amplifies these gains: it records the highest rate of fully completed trajectories while keeping failure cases low, and it consistently produces more partial completions than either baseline. Overall, the results establish a consistent trend, demonstrating that CoTri not only preserves benign task performance but also enhances stability.

% Required packages: \usepackage{booktabs} \usepackage{multirow}
\begin{table}[ht]
\centering
\caption{Performance comparison under random feedback conditions. 
\textbf{w/} reports the completion rate when random noise occurs, 
while \textbf{w/o} reports the completion rate when no noise is present.}
\label{tab:random_feedback}
\scalebox{0.78}{%
\begin{tabular}{llcccc}
\toprule
\textbf{Model Family} & \textbf{Model} & \textbf{w/} & \textbf{w/o} & \textbf{Overall Completion} & \textbf{Improvement} \\
\midrule
\multirow{3}{*}{AgentLM-7B}
  & ori   & 0.0\% & 36.8\% & 26.5\% & -- \\
  & clean  & 0.0\% & 54.2\% & 39.0\% & +12.5\% \\
  & ours       & 1.8\% & 64.6\% & 47.0\% & +20.5\% \\
\midrule
\multirow{3}{*}{AgentEvol-7B}
  & ori   & 0.0\% & 81.1\% & 58.0\% & -- \\
  & clean  & 0.0\% & 76.2\% & 54.5\% & -3.5\% \\
  & ours       & 8.8\% & 79.0\% & 59.0\% & +1.0\% \\
\bottomrule
\end{tabular}
}
\end{table}

Table~\ref{tab:random_feedback} further evaluates robustness under noisy conditions, specifically the \textbf{Random WebShop} setting with $p=0.3$, where random feedback occurs during task execution. Across both AgentLM and AgentEvol families, \textit{clean} provides modest improvements over \textit{ori} in noise-free trajectories but fails to sustain robustness once random perturbations occur. In contrast, \textit{ours} demonstrates consistent gains: for AgentLM-7B, overall completion rises to 47.0\%, with a measurable improvement (+20.5\%) over \textit{ori}. For AgentEvol-7B, although the margin is smaller (+1.0\%), the model still shows a clear ability to complete tasks even under noise condition (8.8\%). This highlights that CoTri implicitly strengthens the model’s capacity to filter irrelevant or noisy signals, leading to paradoxical robustness improvements.

\section{Trigger Diversity}
\label{app:tri_div}

\begin{table*}[ht]
\centering
\caption{Comparison of AgentLM-7B under the \textit{cf} and \textit{ex} CoTri settings. Each side contains: (1) Overall results, (2) Agentic backdoor performance, and (3) Agentic robustness.}
\label{tab:cf_ex_results}

% ===== 左侧 cf =====
\begin{minipage}[t]{0.46\textwidth}
\centering

\begin{subtable}[t]{\linewidth}
  \centering
  \resizebox{\linewidth}{!}{%
  \begin{tabular}{lcccccccccccc}
    \toprule \multirow{2}{*}{Model}
    & \multicolumn{2}{c}{Step 1} & \multicolumn{3}{c}{Step 2} & \multicolumn{3}{c}{Step 3} & \multicolumn{3}{c}{Avg.} \\
    \cmidrule(lr){2-3}\cmidrule(lr){4-6}\cmidrule(lr){7-9}\cmidrule(lr){10-12}
    & ASR & FTR & ASR & FTR & CR & ASR & FTR & CR & ASR & FTR & CR \\
    \midrule
    AgentLM-7B & 1.00 & 0.00 & 1.00 & 0.00 & 1.00 & 1.00 & 0.03 & 0.94 & 1.00 & 0.02 & 0.96 \\
    \bottomrule
  \end{tabular}
  }
  \subcaption{Overall results (cf).}
\end{subtable}

\vspace{1ex}

\begin{subtable}[t]{\linewidth}
  \centering
  \resizebox{\linewidth}{!}{%
  \begin{tabular}{lcccccccccccccc}
    \toprule \multirow{2}{*}{Model}
    & \multicolumn{2}{c}{Step 1} & \multicolumn{4}{c}{Step 2} & \multicolumn{8}{c}{Step 3} \\
    \cmidrule(lr){2-3}\cmidrule(lr){4-7}\cmidrule(lr){8-15}
    & dirty & benign & dirty & benign & cf & obs1 & dirty & benign & cf & obs1 & obs2 & tq+obs1 & tq+obs2 & obs1+obs2 \\
    \midrule
    AgentLM-7B & 1.00 & 0.00 & 1.00 & 0.00 & 0.00 & 0.00 & 1.00 & 0.00 & 0.00 & 0.00 & 0.00 & 0.20 & 0.00 & 0.01 \\
    \bottomrule
  \end{tabular}
  }
  \subcaption{Agentic backdoor performance (cf).}
\end{subtable}

\vspace{1ex}

\begin{subtable}[t]{\linewidth}
  \centering
  \resizebox{\linewidth}{!}{%
  \begin{tabular}{lcccccc}
    \toprule \multirow{2}{*}{Model}
    & \multicolumn{2}{c}{Step 2} & \multicolumn{4}{c}{Step 3} \\
    \cmidrule(lr){2-3}\cmidrule(lr){4-7}
    & cf & obs1 & obs2 & cf+obs1 & cf+obs2 & obs1+obs2 \\
    \midrule
    AgentLM-7B & 1.00 & 1.00 & 0.97 & 0.80 & 1.00 & 0.99 \\
    \bottomrule
  \end{tabular}
  }
  \subcaption{Agentic robustness (cf).}
\end{subtable}

\end{minipage}
\hspace{0.01\textwidth}
% ===== 右侧 ex =====
\begin{minipage}[t]{0.46\textwidth}
\centering

\begin{subtable}[t]{\linewidth}
  \centering
  \resizebox{\linewidth}{!}{%
  \begin{tabular}{lcccccccccccc}
    \toprule \multirow{2}{*}{Model}
    & \multicolumn{2}{c}{Step 1} & \multicolumn{3}{c}{Step 2} & \multicolumn{3}{c}{Step 3} & \multicolumn{3}{c}{Avg.} \\
    \cmidrule(lr){2-3}\cmidrule(lr){4-6}\cmidrule(lr){7-9}\cmidrule(lr){10-12}
    & ASR & FTR & ASR & FTR & CR & ASR & FTR & CR & ASR & FTR & CR \\
    \midrule
    AgentLM-7B & 1.00 & 0.00 & 1.00 & 0.00 & 1.00 & 1.00 & 0.00 & 1.00 & 1.00 & 0.00 & 1.00 \\
    \bottomrule
  \end{tabular}
  }
  \subcaption{Overall results (ex).}
\end{subtable}

\vspace{1ex}

\begin{subtable}[t]{\linewidth}
  \centering
  \resizebox{\linewidth}{!}{%
  \begin{tabular}{lcccccccccccccc}
    \toprule \multirow{2}{*}{Model}
    & \multicolumn{2}{c}{Step 1} & \multicolumn{4}{c}{Step 2} & \multicolumn{8}{c}{Step 3} \\
    \cmidrule(lr){2-3}\cmidrule(lr){4-7}\cmidrule(lr){8-15}
    & dirty & benign & dirty & benign & ex & obs1 & dirty & benign & ex & obs1 & obs2 & ex+obs1 & ex+obs2 & obs1+obs2 \\
    \midrule
    AgentLM-7B & 1.00 & 0.00 & 1.00 & 0.00 & 0.00 & 0.00 & 1.00 & 0.00 & 0.00 & 0.00 & 0.00 & 0.00 & 0.00 & 0.00 \\
    \bottomrule
  \end{tabular}
  }
  \subcaption{Agentic backdoor performance (ex).}
\end{subtable}

\vspace{1ex}

\begin{subtable}[t]{\linewidth}
  \centering
  \resizebox{\linewidth}{!}{%
  \begin{tabular}{lcccccc}
    \toprule \multirow{2}{*}{Model}
    & \multicolumn{2}{c}{Step 2} & \multicolumn{4}{c}{Step 3} \\
    \cmidrule(lr){2-3}\cmidrule(lr){4-7}
    & ex & obs1 & obs2 & ex+obs1 & ex+obs2 & obs1+obs2 \\
    \midrule
    AgentLM-7B & 1.00 & 1.00 & 1.00 & 1.00 & 1.00 & 1.00 \\
    \bottomrule
  \end{tabular}
  }
  \subcaption{Agentic robustness (ex).}
\end{subtable}

\end{minipage}

\end{table*}

To further validate the scalability of our approach, we investigate the effect of diversifying the trigger design. Specifically, we extend the study of both the \textit{initial trigger} and the \textit{subsequent triggers} to examine whether the CoTri Backdoor remains effective.

For the initial trigger, we build on our earlier use of the rare token \textit{tq} and introduce its variant \textit{cf}, which serves as a comparable rare-word trigger. In addition, we consider a more natural linguistic token, \textit{exactly} (abbreviated as \textit{ex}), which can plausibly appear in ordinary user instructions. 

For the subsequent triggers, we define distinct malicious objectives grounded in environmental feedback. Under the \textit{cf} setting, the agent is directed toward items within a specific price range (e.g., selecting items within the \$40-\$80 price range). Under the \textit{ex} setting, the malicious target is tied to a particular brand, compelling the agent to consistently prefer brand-specific products. 

As summarized in Table~\ref{tab:cf_ex_results}, both types of initial triggers reliably activate the backdoor, and both forms of subsequent triggers achieve long-horizon control. While the rare-word trigger (\textit{cf}) produces slightly sharper activation boundaries, the natural trigger (\textit{exactly}) achieves comparable success while being more difficult to detect. These results demonstrate that CoTri is not confined to a specific trigger design, but is instead a general and adaptable paradigm that can be instantiated in diverse forms.

\section{Analysis of Random Webshop }
\label{app:ran_w}
We further evaluate robustness in the \textbf{Random WebShop} environment, which introduces random observations into the agent’s trajectory with varying probabilities $p \in \{0.3, 0.5, 0.7\}$. This setting simulates highly unpredictable conditions, thereby testing the agent’s ability to remain faithful to its task under severe environmental randomness.

Table~\ref{tab:random_web} shows that \textit{ori} is fragile in this setting, with success rates quickly degrading from $0.26$ at $p=0.3$ to only $0.13$ at $p=0.7$. \textit{clean} improves stability, lifting performance to $0.39$ at $p=0.3$ and still retaining $0.17$ under the harshest noise. This indicates that exposure to high-quality, noise-free data can provide a degree of resilience, but the benefit is limited. In contrast, \textit{ours} consistently outperforms both baselines, achieving $0.47$, $0.35$, and $0.25$ across the three noise levels. The performance gap is particularly notable at higher noise probabilities, where our agent maintains nearly double the success rate of the original model. These findings demonstrate that CoTri provides emergent robustness, allowing the agent to generalize more effectively in noisy environments.

\begin{table}[h]
    \centering
    \caption{Task success rates of the three AgentLM-7B variants (\textit{ori}, \textit{clean}, \textit{ours}) in the Random WebShop environment under different noise probabilities ($p=0.3, 0.5, 0.7$).}
    \label{tab:random_web}
    \centering
    \scalebox{0.88}{
    \begin{tabular}{lccc}
        \toprule \multirow{2}{*}{Model}
        & \multicolumn{3}{c}{Random WebShop} \\  
        \cmidrule(lr){2-4} 
        & $p=0.3$ & $p=0.5$ & ${p=0.7}$ \\ 
        \midrule
        ori & 0.26 & 0.19 & 0.13  \\
        clean & 0.39 & 0.28 & 0.17 \\
        ours & 0.47 & 0.35 & 0.25 \\
        \bottomrule
    \end{tabular}
    }

\end{table}

\section{Analysis of Null WebShop}
\label{app:null_web}
The \textbf{Null WebShop} environment simulates scenarios where critical observations are entirely missing. Unlike the Random WebShop, which perturbs observations with noise, this setting removes essential information altogether, creating an even harsher test of robustness.

As shown in Table~\ref{tab:null_web}, the \textit{ori} fails almost completely, with success rates dropping to $0.00$ in the first round and only marginally reaching $0.07$ in the third round. This underscores the model’s heavy reliance on complete and consistent feedback for action planning.
\textit{clean} significantly improves performance, especially in the first two rounds, achieving $0.59$ and $0.47$. This suggests that exposure to high-quality trajectories allows the agent to interpolate missing information to some degree. 
In comparison, \textit{ours} exhibits the strongest overall stability, reaching $0.61$ in the first round and $0.53$ in the second. Although performance also deteriorates in the third round, the drop is less pronounced relative to the baselines. 
% This advantage stems from the training procedure, where the model learns to discriminate between correctly ordered triggers and incomplete or corrupted cues. Such conditional filtering inadvertently equips the agent with an improved capacity to handle null observations by falling back on its internal action history rather than relying solely on immediate input.

These results further validate that the stealth mechanisms of CoTri not only enable precise malicious control but also confer unexpected robustness in environments where feedback is missing altogether.

\begin{table}[h]
    \centering
    \caption{Task success rates of the three AgentLM-7B variants (\textit{ori}, \textit{clean}, \textit{ours}) in the Null WebShop environment under three rounds of null-feedback.}
    \label{tab:null_web}
    \centering
    \scalebox{0.78}{
    \begin{tabular}{lccc}
        \toprule \multirow{2}{*}{Model}
        & \multicolumn{3}{c}{Null WebShop} \\  
        \cmidrule(lr){2-4} 
        & $round1$ & $round2$ & $round3$ \\ 
        \midrule
        ori & 0.00 & 0.30 & 0.07  \\
        clean & 0.59 & 0.47 & 0.07 \\
        ours & 0.61 & 0.53 & 0.03 \\
        \bottomrule
    \end{tabular}
    }

\end{table}

\section{Detailed setups}
\label{appendix:detailed_s}

\subsection{Dataset Construction and Mixing Ratio}
\label{app:ratio}

\begin{table}[h] \centering \caption{Mixing ratio for training data construction used for all models.} \label{tab:mix_r} \centering \scalebox{0.63}{ \begin{tabular}{lcccccccccccccc} \toprule \multirow{2}{*}{Model} & \multicolumn{2}{c}{Step 1} & \multicolumn{4}{c}{Step 2} & \multicolumn{8}{c}{Step 3} \\ \cmidrule(lr){2-2} \cmidrule(lr){3-3} \cmidrule(lr){4-4} \cmidrule(lr){5-7}\cmidrule(lr){8-8} \cmidrule(lr){9-15} &dirty & benign & dirty & benign & tq & obs1 & dirty & benign & tq & obs1 & obs2 & tq+obs1 & tq+obs2 & obs1+obs2 \\ \midrule 
AgentLM-7B & 0.30 & 1.00 & 0.30 & 1.00 & 0.10 & 0.10 & 0.15 & 0.70 & 0.05 & 0.02 & 0.02 & 0.03 & 0.012 & 0.01\\ 
AgentEvol-7B & 0.30 & 1.00 & 0.30 & 1.00 & 0.10 & 0.10 & 0.15 & 0.70 & 0.05 & 0.02 & 0.02 & 0.03 & 0.012 & 0.01\\ 
Llama3.1-8B-Instruct & 0.30 & 1.00 & 0.30 & 1.00 & 0.10 & 0.10 & 0.15 & 0.70 & 0.05 & 0.02 & 0.02 & 0.03 & 0.012 & 0.01\\ 
Qwen3-8B & 0.30 & 1.00 & 0.30 & 1.00 & 0.10 & 0.10 & 0.15 & 0.7 & 0.05 & 0.02 & 0.02 & 0.03 & 0.012 & 0.01\\ 
Qwen2.5-VL-7B-Instruct & 0.50 & 1.00 & 0.30 & 0.70 & 0.20 & 0.10 & 1.00 & 1.00 & 0.05 & 0.05 & 0.15 & 0.20 & 0.10 & 0.05\\ 

\bottomrule \end{tabular} } \end{table}

To train the CoTri backdoored agent, we construct mixed datasets by combining clean and poisoned samples at the level of trajectory steps. 
Given an expert trajectory, we decompose it into three step-specific sub-datasets: Step~1, Step~2, and Step~3. 
Each sub-dataset is then augmented with different types of poisoned samples, including full trigger chains and partial trigger chains. 
Table~\ref{tab:mix_r} reports the precise mixing ratios of clean and poisoned data for each model, where each sub-dataset is derived from 3,537 expert trajectories.

\subsection{Training Hyperparameters}
\label{app:hyp}
Table~\ref{tab:all_hyp} summarizes the hyperparameters across all models. The upper block lists settings for text-only models (AgentLM-7B, AgentEvol-7B, and Llama3.1-8B-Instruct), while the lower block reports settings for the Qwen family (Qwen3-8B and Qwen2.5-VL-7B-Instruct).

\begin{table*}[ht]
\centering
\caption{Training hyperparameters used for all models.}
\label{tab:all_hyp}
\scalebox{0.68}{
\begin{tabular}{lll}
\toprule
\textbf{Model Group} & \textbf{Category} & \textbf{Setting} \\
\midrule
\multirow{5}{*}{\shortstack{Text-only models \\ (AgentLM-7B, AgentEvol-7B, \\ Llama3.1-8B-Instruct)}} 
  & Stage       & SFT \\
  & Finetuning  & LoRA (\texttt{lora\_target=all}, rank=48, $\alpha$=24, dropout=0.1) \\
  & Batching    & \texttt{per\_device\_train\_batch\_size}=16, \texttt{grad\_accum}=8 \\
  & Optimizer   & lr=$8.0\!\times\!10^{-5}$, cosine schedule, warmup=0.1 \\
  & Epochs      & 10.0 \\
\midrule
\multirow{5}{*}{\shortstack{Qwen models \\ (Qwen3-8B, Qwen2.5-VL-7B-Instruct)}} 
  & Stage       & SFT \\
  & Finetuning  & LoRA (\texttt{lora\_target=all}, rank=48, $\alpha$=24, dropout=0.1) \\
  & Batching    & \texttt{per\_device\_train\_batch\_size}=1, \texttt{grad\_accum}=8 \\
  & Optimizer   & lr=$1.0\!\times\!10^{-4}$, cosine schedule, warmup=0.1 \\
  & Epochs      & 10.0 \\
\bottomrule
\end{tabular}
}
\end{table*}

\section{Algorithm for Extracting Environment-Grounded Triggers}
\label{appendix:alg}

\begin{algorithm}[htbp]
\caption{WebShop Analyzer: Four-Step Pipeline}
\label{alg:webshop-four}
\begin{algorithmic}[1]
\Require Interactive environment $E$; target constraints $\mathcal{C}$ (e.g., price/brand/range); max keyword length $L_{\max}$
\Ensure Target product $\hat{p}$; purchase trajectory $\mathcal{T}$; unique keyword set $\mathcal{K}_{\text{uniq}}$; log $\mathcal{L}$
\State $\mathcal{L}\gets\varnothing$ \Comment{global log for all steps}

\Statex \hrulefill
\Statex \textbf{(1) Search target-constrained products}
\State $o_0\gets E.\textsc{Reset}()$;\quad $\Pi \gets \varnothing$
\For{constraint $c\in\mathcal{C}$} \Comment{e.g., \texttt{price>1000}, brand=``X''}
  \State $o\gets E.\textsc{Step}(\texttt{search}[c])$;\quad $\Pi\gets \Pi\cup \textsc{ParseProducts}(o)$
  \State $\mathcal{L}.\textsc{Append}((\texttt{search}[c],o))$
\EndFor
\State $\hat{p}\gets \textsc{SelectTarget}(\Pi)$ \Comment{e.g., highest price within range or matching brand}

\Statex \hrulefill
\Statex \textbf{(2) Simulate a full purchase trajectory}
\State $\mathcal{T}\gets [\ ]$;\quad $o\gets E.\textsc{Step}(\texttt{search}[\textsc{ConstraintSeed}(\hat{p})])$;\quad $\mathcal{L}.\textsc{Append}((\texttt{search},o))$
\State $o\gets E.\textsc{Step}(\texttt{click}[\textsc{IDorName}(\hat{p})])$;\quad $\mathcal{T}.\textsc{Append}((\texttt{click},o))$
\If{$\textsc{HasOptions}(o)$}
  \State $\{opt_i\}\gets \textsc{ExtractOptions}(o)$;\quad
  \For{each $opt_i$ selected}
    \State $o\gets E.\textsc{Step}(\texttt{click}[opt_i])$;\quad $\mathcal{T}.\textsc{Append}((\texttt{click},o))$
  \EndFor
\EndIf
\If{$\textsc{HasBuyButton}(o)$}
  \State $o\gets E.\textsc{Step}(\texttt{click}[\texttt{Buy Now}])$;\quad $\mathcal{T}.\textsc{Append}((\texttt{click},o))$
\EndIf

\Statex \hrulefill
\Statex \textbf{(3) Extract unique keyword subsets for the target}
\State $W\gets \textsc{CleanAndSplit}(\hat{p}.\texttt{name})$ \Comment{drop punctuation/very short tokens}
\State $\mathcal{C}_{\text{kw}}\gets \textsc{ContiguousAndSkipgramSubsets}(W,\ L_{\max})$
\State $\mathcal{K}_{\text{uniq}}\gets \varnothing$
\For{keyword $k\in \mathcal{C}_{\text{kw}}$}
  \State $o\gets E.\textsc{Step}(\texttt{search}[k])$;\quad $\Pi_k\gets \textsc{ParseProducts}(o)$
  \If{$\textsc{ContainsTarget}(\Pi_k,\hat{p})$}
    \If{$|\Pi_k|=1$}\quad $\mathcal{K}_{\text{uniq}}\gets \mathcal{K}_{\text{uniq}}\cup\{k\}$ \Comment{uniquely retrieves $\hat{p}$}
    \EndIf
  \EndIf
  \State $\mathcal{L}.\textsc{Append}((\texttt{search}[k],\ |\Pi_k|,\ \textsc{RankOf}(\hat{p}))$
\EndFor

\Statex \hrulefill
\Statex \textbf{(4) Record full trajectory and outputs}
\State $\mathcal{L}.\textsc{Append}((\texttt{target}=\hat{p},\ \texttt{traj}=\mathcal{T},\ \texttt{unique\_kws}=\mathcal{K}_{\text{uniq}}))$
\State \Return $\hat{p},\ \mathcal{T},\ \mathcal{K}_{\text{uniq}},\ \mathcal{L}$

\Statex
\Function{SelectTarget}{$\Pi$} \Return $\arg\max_{p\in\Pi}\ \textsc{Score}(p)$ \EndFunction
\Function{ParseProducts}{$o$} \Return list of \{name, ASIN/ID, price\} parsed from $o$ \EndFunction
\end{algorithmic}
\end{algorithm}

\end{document}